\pdfoutput=1

\documentclass[11pt]{article}

\usepackage[]{acl}

\usepackage{times}
\usepackage{latexsym}

\usepackage[T1]{fontenc}

\usepackage[utf8]{inputenc}

\usepackage{microtype}

\usepackage{amsmath,amsfonts,bm}

\def\eqref#1{equation~\ref{#1}}

\def\1{\bm{1}}

\DeclareMathAlphabet{\mathsfit}{\encodingdefault}{\sfdefault}{m}{sl}
\SetMathAlphabet{\mathsfit}{bold}{\encodingdefault}{\sfdefault}{bx}{n}

\usepackage{tikz-dependency}
\usepackage{booktabs}
\usepackage{comment}
\usepackage{graphicx} 
\usepackage{amssymb}
\usepackage{times}
\usepackage{latexsym}
\usepackage{pifont}
\usepackage{times}
\usepackage{latexsym}
\usepackage{comment}

\usepackage{url}
\usepackage{amsmath}

\usepackage{CJKutf8}
\usepackage[boxed]{algorithm}
\usepackage{caption}
\usepackage{algpseudocode}

\usepackage[T1]{fontenc}
\usepackage[utf8]{inputenc}

\usepackage{microtype}
\usepackage{tcolorbox}
\usepackage{tikz,pgfplots}
\usepgfplotslibrary{groupplots}
\usepackage{nicefrac} 
\pgfplotsset{compat=1.13}

\usepackage{enumerate}
\usepackage{amssymb}

\definecolor{g-red}{HTML}{DB4437}
\definecolor{g-blue}{HTML}{4285F4}
\definecolor{g-green}{HTML}{0F9D58}
\definecolor{g-yellow}{HTML}{F4B400}
\definecolor{g-orange}{HTML}{FF9800}
\definecolor{g-grey}{HTML}{9E9E9E}
\definecolor{shannon}{HTML}{304FFE}
\definecolor{uw}{RGB}{138,43,226}
\definecolor{stanford}{RGB}{255,69,0}
\definecolor{const}{RGB}{68, 110, 182}
\definecolor{head}{RGB}{246, 180, 32}
\definecolor{freq}{RGB}{0, 0, 0}
\definecolor{ao}{rgb}{0.0, 0.5, 0.0}
\definecolor{asparagus}{rgb}{0.53, 0.66, 0.42}
\definecolor{amber}{rgb}{1.0, 0.49, 0.0}
\definecolor{alizarin}{rgb}{0.82, 0.1, 0.26}
\definecolor{applegreen}{rgb}{0.55, 0.71, 0.0}
\definecolor{amethyst}{rgb}{0.6, 0.4, 0.8}
\definecolor{auburn}{rgb}{0.43, 0.21, 0.1}

\newenvironment{tightitemize}%
  {\begin{itemize} %
    \setlength{\itemsep}{0.5pt}%
    \setlength{\parskip}{0.5pt}%
    }%
  {\end{itemize}}

\usepackage{caption}
\usepackage{booktabs}

\usepackage{babel}

\usepackage{array} 

\usepackage{multirow}
\usepackage{subcaption}
\usepackage{booktabs} 
\usepackage{threeparttable}
\usepackage{graphicx}
\usepackage{adjustbox}
\usepackage{multicol}
\usepackage{tablefootnote} 
\usepackage[referable]{threeparttablex}
\title{Instruction Tuning for Large Language Models: A Survey}

\author{Shengyu Zhang$^{\spadesuit}$, Linfeng Dong$^{\spadesuit}$, Xiaoya Li$^{\blacktriangledown}$, Sen Zhang$^{\spadesuit}$ \\  
{\bf Xiaofei Sun$^{\spadesuit}$, Shuhe Wang$^{\clubsuit}$, Jiwei Li$^{\spadesuit}$, Runyi Hu$^{\spadesuit}$ }\\
{\bf Tianwei Zhang$^\blacktriangle$, Fei Wu$^{\spadesuit}$ and Guoyin Wang$^{\blacklozenge}$ }}

\begin{document}
\maketitle
\begin{abstract}
This paper surveys  research works in the quickly advancing field of instruction tuning (IT), which can also be referred to as supervised fine-tuning (SFT)\footnote{In this paper, unless specified otherwise, supervised fine-tuning (SFT) and instruction tuning (IT) are used interchangeably. }, 
a crucial technique to enhance the capabilities and controllability of large language models (LLMs).
Instruction tuning 
refers to the process of  further training LLMs on a dataset consisting   
of \textsc{(instruction, output)} pairs
 in a supervised fashion, 
which  bridges the gap between the next-word prediction objective of LLMs and  the users' objective of having LLMs adhere to human instructions.
In this work, we make a systematic review of the literature, including  the general methodology of SFT, 
the construction of SFT datasets, the training of SFT models, 
and applications to different modalities, domains and application, along with analysis on aspects that influence the outcome of SFT (e.g., generation of instruction outputs, size of the instruction dataset, etc). We also 
review the potential pitfalls of SFT along with criticism against it, along with efforts
pointing out current deficiencies of existing strategies and suggest some avenues for fruitful research.
\let\thefootnote\relax\footnotetext{$^{\spadesuit}$Zhejiang University, $^{\clubsuit}$Shannon.AI, $^\blacktriangle$Nanyang Technological University, $^{\blacklozenge}$Amazon, $^\blacktriangledown$University of Washington}
\let\thefootnote\relax\footnotetext{Email: sy\_zhang@zju.edu.cn}
\let\thefootnote\relax\footnotetext{Project page can be found at: \url{https://github.com/xiaoya-li/Instruction-Tuning-Survey}}
\let\thefootnote\relax\footnotetext{\bf * The latest update was on Aug. 11, 2025 (Version 6).}
\addtocounter{footnote}{0}\let\thefootnote\svthefootnote
\end{abstract}

\section{Introduction}

The field of large language models (LLMs) has witnessed remarkable progress in recent years. LLMs such as GPT-3~\citep{Brown2020LanguageMA}, PaLM~\citep{Chowdhery2022PaLMSL}, and LLaMA~\citep{Touvron2023LLaMAOA} have demonstrated impressive capabilities across a wide range of natural language tasks~\citep{Zhao2021CalibrateBU, Wang2022SelfConsistencyIC, wang2023gpt, wan2023gpt, sun2023text, wei2023zero, Li2023EvaluatingCI, Gao2023ExploringTF, Yao2023TreeOT, yang2022re3, qian2022controllable,  Lee2022CoAuthorDA, Yang2022TailorAP, gao2023enabling, Ning2023AlbumSW, liu2021makes, wiegreffe2021reframing, sun2023pushing, Sun2023AnsweringAQ, Adlakha2023EvaluatingCA, Chen2023DistinguishBA}. 
One of the major issues with LLMs is the mismatch between the training objective and users' objective: 
LLMs are typically trained on minimizing 
the
contextual word prediction error on large corpora;
while users want  the model to  "follow their instructions helpfully and safely"~\citep{radford2019language, brown2020language, Fedus2021SwitchTS, rae2021scaling, thoppilan2022lamda}

To address this mismatch, instruction tuning (IT), which can also be referred to as supervised fine-tuning (SFT), is proposed, serving as an
 effective technique to enhance the capabilities and controllability of large language models. It involves further training LLMs using 
\textsc{(instruction, output)} pairs, where  \textsc{instruction} denotes the human instruction for the model, and  \textsc{output} denotes the desired output that follows the \textsc{instruction}. 
The benefits of SFT are threefold: 
(1) Finetuning an LLM on the instruction dataset bridges the gap between the next-word prediction objective of LLMs and  the users' objective of instruction following; 
(2) SFT allows for a more controllable and predictable model behavior compared to standard LLMs. The instructions serve to constrain the model's outputs to align with the desired response characteristics or domain knowledge, providing a channel for humans to intervene with the model's behaviors; and
(3) SFT is computationally efficient and can help LLMs rapidly adapt  to a specific domain without extensive retraining or architectural changes.

Despite its effectiveness, SFT  also poses challenges: (1) Crafting high-quality instructions that properly cover the desired target behaviors is non-trivial:
existing instruction datasets are usually limited in quantity, diversity, and creativity;
(2) there has been an increasing concern that SFT only  improves on  tasks that are heavily supported in the SFT training dataset \cite{gudibande2023false}; and
(3) there has been an intense criticism that SFT only
 captures surface-level patterns and styles (e.g., the output format) rather than comprehending and learning the  task \cite{Kung2023DoMR}.
Improving instruction adherence and handling unanticipated model responses remain open research problems. These challenges highlight the importance of further investigations, analysis, and summarization in this field, to optimize the fine-tuning process and better understand the behavior of instruction tuned LLMs.

In the literature, there has been an increasing research interest in analysis and discussions on LLMs, including pre-training methods \cite{zhao2023survey}, 
reasoning abilities \cite{huang2022towards},
downstream applications \cite{yang2023harnessing,sun2023pushing}, but rarely on the topic of LLM instruction tuning.
This survey attempts to 
fill this blank, 
organizing  the most up-to-date state of knowledge on this quickly advancing field.
Specifically, 
\begin{tightitemize}
\item Section \ref{sec:Methodology} presents the general methodology employed in instruction tuning.
\item Section \ref{sec:Datasets} outlines the construction process of commonly-used SFT representative datasets, along with multi-step reasoning datasets designed to enhance LLM performance on complex reasoning tasks such as mathematics and coding.
\item Section \ref{sec:Instruction_fine-tuned_LLMs} presents representative instruction tuned  models.
\item Section \ref{sec:Multi-modality_Instruction_Fine-tuning} reviews multi-modality techniques and datasets for instruction tuning, including images, speech, and video.
\item Section \ref{application} reviews efforts to adapt LLMs to different domains and applications using the SFT strategy.
\item Section \ref{Efficient_fine-tuning_techniques} reviews explorations to make instruction tuning more efficient, reducing the computational and time costs associated with adapting large models.
\item Section \ref{analysis} presents the evaluation of SFT models, analysis on them, along with criticism against them. 
\item Section \ref{sec:the_role_of_instruction_fine_tuning} analyzes the role of SFT in comparison with recent, highly effective reinforcement learning–based methods (e.g., RLHF, DPO, and GRPO).
\end{tightitemize}

\begin{figure*}
    \centering
    \includegraphics[width=\linewidth]{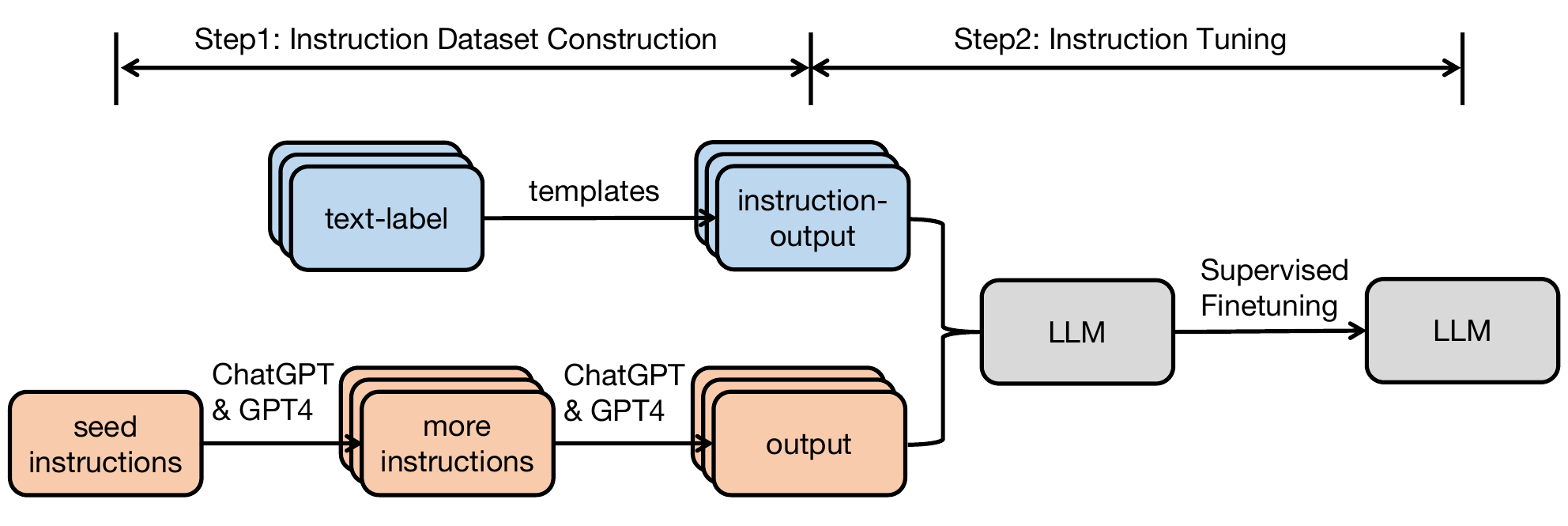}   
    \caption{General pipeline of instruction tuning.}
    \label{fig:method-overall}
\end{figure*}

\section{Methodology} \label{sec:Methodology}
In this section, we describe the general pipeline employed in instruction tuning.
\subsection{Instruction Dataset Construction}
Each instance in an instruction dataset consists of three elements: an instruction, which is a natural language text sequence to specify the task (e.g., {\it write a thank-you letter to XX for XX}, {\it write a blog on the topic of XX}, etc); an optional input which provides supplementary information for context; and an anticipated output based on the instruction and the input. 

There are generally two methods for constructing instruction datasets:
\begin{tightitemize}
\item Data integration from annotated natural language datasets. In this approach, (instruction, output) pairs are collected from existing annotated natural language datasets by using templates to transform text-label pairs to (instruction, output) pairs. Datasets such as Flan~\citep{longpre2023flan} and P3~\citep{sanh2021multitask} are constructed based on the data integration strategy.

\item Generating outputs using LLMs: An alternate way to quickly gather the desired outputs to given instructions is to employ LLMs such as GPT-3.5-Turbo or GPT4 instead of manually collecting the outputs. Instructions can come from two sources: (1) manually collected; or (2) expanded based a small handwritten seed instructions using LLMs. Next, the collected instructions are fed to LLMs to obtain outputs. Datasets such as InstructWild~\citep{instructionwild} and Self-Instruct~\citep{wang2022self} are geneated following this approach.
\end{tightitemize}

For multi-turn conversational SFT datasets, we can have  large language models self-play different roles (user and AI assistant) to generate messages in a conversational format ~\citep{xu2023baize}.

\subsection{Instruction Tuning / Supervised Fine-tuning}
Based on the collected SFT dataset, a pretrained model can be directly fune-tuned in a fully-supervised manner, where given the instruction and the input, the model is trained by predicting each token in the output sequentially. 

\section{Datasets}  
\label{sec:Datasets}
In this section, we detail instruction tuning datasets in the community, categorizing them into three classes: (1) Human-crafted Data, (2) Synthetic Data via Distillation, and (3) Synthetic Data via Self-improvement. 
Further more, in light of the impressive performance of recent multi-step reasoning LLMs (e.g., OpenAI o1 \cite{jaech2024openai}, DeepSeek-R1 \cite{guo2025deepseek}), this section also presents a detailed overview of how reasoning datasets are constructed. These datasets, typically built using one or a combination of the three strategies mentioned above, are specifically designed to enhance LLMs’ multi-step thinking capabilities.
Below, we describe some widely-used datasets, and for full collected datasets we put them in Appendix \ref{appendix:datasets}.

\subsection{Human-crafted Data}

Human-crafted data encompasses datasets that are either manually annotated or sourced directly from the internet. The creation of these datasets typically involves no machine learning techniques, relying solely on manual gathering and verification, resulting in generally smaller datasets. Below are some widely-used human-crafted datasets:

\subsubsection{Natural Instructions}
\begin{figure}[t]
  \centering
  \begin{minipage}[t]{0.48\textwidth}
    \centering
    \includegraphics[width=1\textwidth]{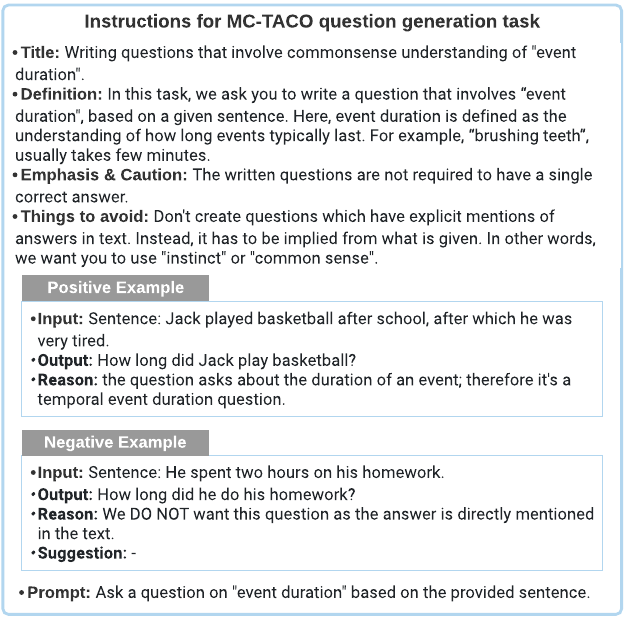}
    \subcaption{An example of \textsc{instructions} in Natural Instruction dataset. }
  \end{minipage}%
  \hfill
  \begin{minipage}[t]{0.48\textwidth}
    \centering
    \includegraphics[width=1\textwidth]{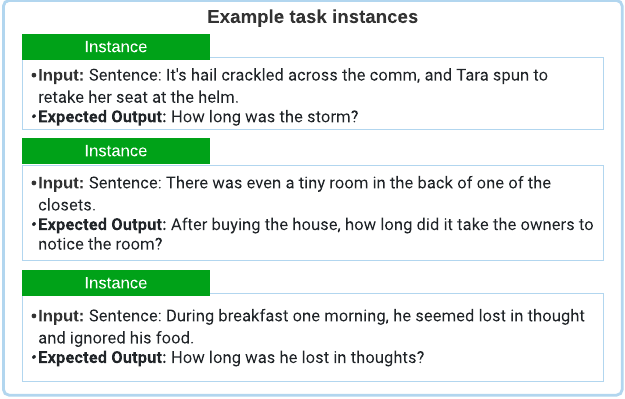}
    \subcaption{An example of \textsc{instances} in Natural Instruction dataset.}
  \end{minipage}
  \caption{The figure is adapted from \citet{mishra2021cross}.}
  \label{fig:data_natural_instruction}
\end{figure}
Natural Instructions~\citep{mishra2021cross} is a human-crafted English instruction dataset consisting of 193K  instances, coming from 61 distinct NLP tasks. 
The dataset is comprised of "instructions" and "instances". Each instance in the "instructions" is a task description consisting of 7 components: title, definition, things to avoid emphasis/caution, prompt, positive example, and negative example. Subfigure (a) in Figure~\ref{fig:data_natural_instruction}  gives an example of the "instructions". 
"Instances" consists of ("input", "output") pairs, which are the input data and textual result that follows the given instruction correctly. 
Subfigure (b) in Figure~\ref{fig:data_natural_instruction} gives an example of the instances. 

The  data  comes from existing NLP datasets of 61 tasks. The authors collected the "instructions" by  referring to the dataset annotating instruction file. Next, the authors constructed the "instances" by unifying data instances across all NLP datasets to ("input", "output") pairs. 

\subsubsection{P3}
P3 (Public Pool of Prompts)~\citep{sanh2021multitask} is an instruction tuning dataset constructed by integrating 170 English NLP datasets and 2,052 English prompts. Prompts, which are sometimes named \textit{task templates}, are functions that map a data instance in a conventional NLP task (e.g., question answering, text classification) to a natural language input-output pair.

Each instance in P3 has
three components: 
 "inputs", "answer\_choices", and “targets". "Inputs" is a sequence of text that describes the task in natural language (e.g., \textit{"If he like Mary is true, is it also true that he like Mary's cat?"}). 
"Answer choices" is a list of text string that are applicable responses to the given task (e.g., \textit{["yes", "no", "undetermined"]}). "Targets" is a text string that is the correct response to the given "inputs" (e.g., "yes"). 
The authors built PromptSource, a tool for creating high-quality prompts collaboratively and an archive for open-sourcing high-quality prompts. 

The P3 dataset was built by randomly sampling a prompt from multiple prompts in the PromptSource and mapping each instance into a ("inputs", "answer choices", "targets") triplet.

\subsubsection{xP3}
xP3 (Crosslingual Public Pool of Prompts)~\citep{muennighoff2022crosslingual} is a multilingual instruction dataset consisting of 16 diverse natural language tasks in 46 languages. Each instance in the dataset has two components: "inputs" and "targets". "Inputs" is a task description in natural language. "Targets" is the textual result that follows the "inputs" instruction correctly. 

The original data in xP3 comes from three sources: the English instruction dataset P3, 4 English unseen tasks in P3 (e.g., translation, program synthesis), and 30 multilingual NLP datasets. 
The authors built the xP3 dataset by sampling human-written task templates from PromptSource and then filling templates to transform diverse NLP tasks into a unified formalization. For example, a task template for the natural language inference task is as follows: \textit{“If {Premise} is true, is it also true that {Hypothesis}?”}; "yes", "maybe", no" with respect to the original task labels "entailment (0)", "neutral (1)" and "contradiction (2)". 

\subsubsection{Flan 2021}
Flan 2021~\citep{longpre2023flan} is an English instruction dataset constructed by transforming 62 widely-used NLP benchmarks (e.g., SST-2, SNLI, AG News, MultiRC) into language input-output pairs. Each instance in the Flan 2021 has "input" and "target" components. "Input" is a sequence of text that describes a task via a natural language instruction (e.g., \textit{"determine the sentiment of the sentence 'He likes the cat.' is positive or negative?"}). "Target" is a textual result that executes the "input" instruction correctly (e.g., \textit{"positive"}). 
The authors transformed conventional NLP datasets into input-target pairs by: 
Step 1: manually composing instruction and target templates;
Step 2: filling templates with data instances from the dataset. 

\subsubsection{LIMA}
LIMA~\citep{Zhou2023LIMALI} is an English instruction dataset consisting of a train set with 1K data instances and a test set with 300 instances. 
The train set contains 1K ("instruction", "response") pairs. 
For the training data, 
75\%  are sampled from three community question \& answers websites (i.e., Stack Exchange, wikiHow, and the Pushshift Reddit Dataset~\citep{Baumgartner2020ThePR}); 20\%  are manually written by a set of the authors (referred Group A) inspired by their interests; 5\% are sampled from the Super-Natural Instructions dataset~\citep{Wang2022SuperNaturalInstructionsGV}. As for the valid set, the authors sampled 50 instances from the Group A author-written set.  
The test set contains 300 examples, with 76.7\% written by another group (Group B) of authors and 23.3\% sampled from the Pushshift Reddit Dataset~\citep{Baumgartner2020ThePR}, which is a collection of questions \& answers within the Reddit community.

\subsubsection{Super-Natural Instructions}
\begin{figure}[t]
  \centering
  \begin{minipage}[t]{0.48\textwidth}
    \centering
    \includegraphics[width=1\textwidth]{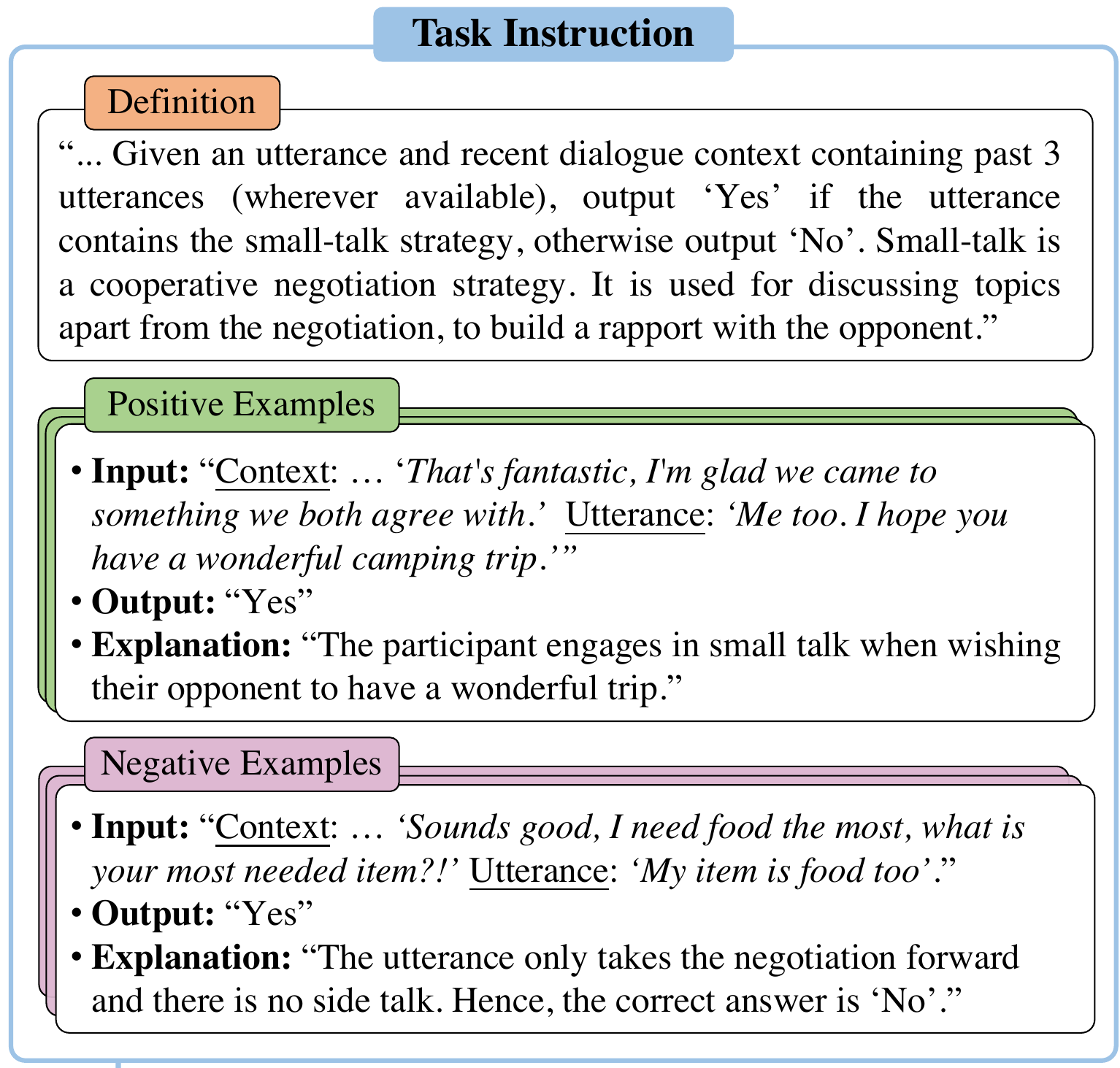}
    \subcaption{An example of \textsc{instructions} in Super-Natural Instruction dataset. }
  \end{minipage}%
  \hfill
  \begin{minipage}[t]{0.48\textwidth}
    \centering
    \includegraphics[width=1\textwidth]{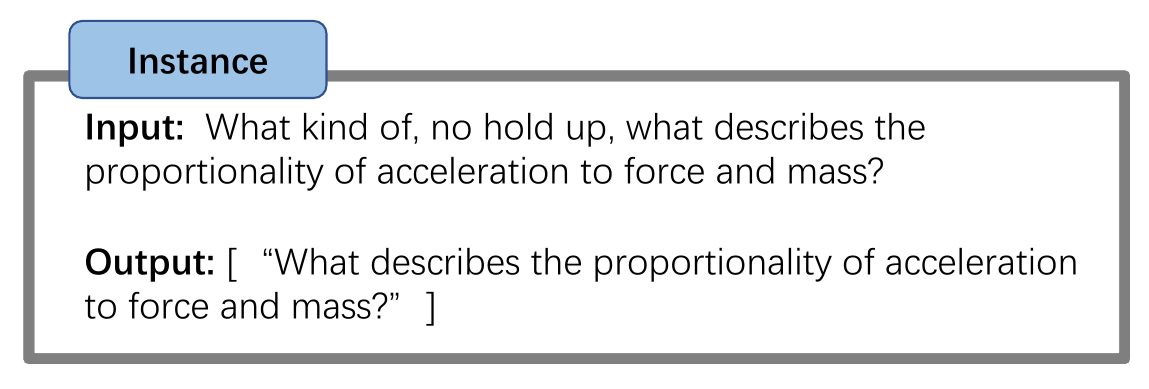}
    \subcaption{An example of \textsc{instances} in Super-Natural Instruction dataset.}
  \end{minipage}
  \caption{The figure is adapted from \citet{wang2022super}.}
  \label{fig:data_supernatural}
\end{figure}
Super Natural Instructions~\citep{supernaturalinstructions} is a multilingual instruction collection composed of 1,616 NLP tasks and 5M task instances, covering 76 distinct task types (e.g., text classification, information extraction, text rewriting, text composition and etc.) and 55 languages. Each task in the dataset consists of an "instruction" and "task instances". Specifically, "instruction" has three components: a "definition" that describes the task in natural language; "positive examples" that are samples of inputs and correct outputs, along with a short explanation for each; and "negative examples" that are samples of inputs and undesired outputs, along with a short explanation for each, as shown in Figure~\ref{fig:data_natural_instruction}  (a). 
 "Task instances" are data instances comprised of textual input and a list of acceptable textual outputs, as shown in Figure~\ref{fig:data_natural_instruction}  (b).
The original data in Super Natural Instructions comes from three sources: (1) existing public NLP datasets (e.g., CommonsenseQA); (2) applicable intermediate annotations that are generated through a crowdsourcing process (e.g., paraphrasing results to a given question during a crowdsourcing QA dataset); (3) synthetic tasks that are transformed from symbolic tasks and rephrased in a few sentences (e.g., algebraic operations like number comparison).

\subsubsection{Dolly}
\begin{table*}[t]
\centering
\small
\scalebox{1.0}{
\begin{tabular}{ll}
\toprule
{\bf Instruction Type}  & {\bf Example}  \\\midrule
Open Q\&A & Why do people like comedy movies?  \\
Closed Q\&A & Does outbreeding or inbreeding benefit the offspring more?\\
Information Extraction & Who was John Moses Browning?\\ 
Information Summarization & Please summarize what Linkedin does.\\
Brainstorming & Give me some ideas to manage my manager. \\
Classification & Identify which animal species is alive or extinct: Palaeophis, Giant Tortoise\\
Creative writing & Write a short story about a person who discovers a hidden room in their house. \\ 
\bottomrule
\end{tabular}
}
\caption{Examples of instructions in Dolly V1~\citep{conover2023free}.}
\label{tab:dolly1_example}
\end{table*}
Dolly~\citep{conover2023free} is an English instruction dataset with 15,000 human-generated data instances designed to enable LLMs to interact with users akin to ChatGPT. 
The dataset is designed for simulating a wide range of human behaviors, covering 7 specific types: open Q\&A, closed Q\&A, extracting information from Wikipedia, summarizing information from Wikipedia, brainstorming, classification, and creative writing. Examples of each task type in the dataset are shown in Table~\ref{tab:dolly1_example}. 

\subsubsection{OpenAssistant Conversations}
\begin{figure}[t]
  \centering
  \begin{minipage}[t]{0.5\textwidth}
    \centering
    \includegraphics[width=1\textwidth]{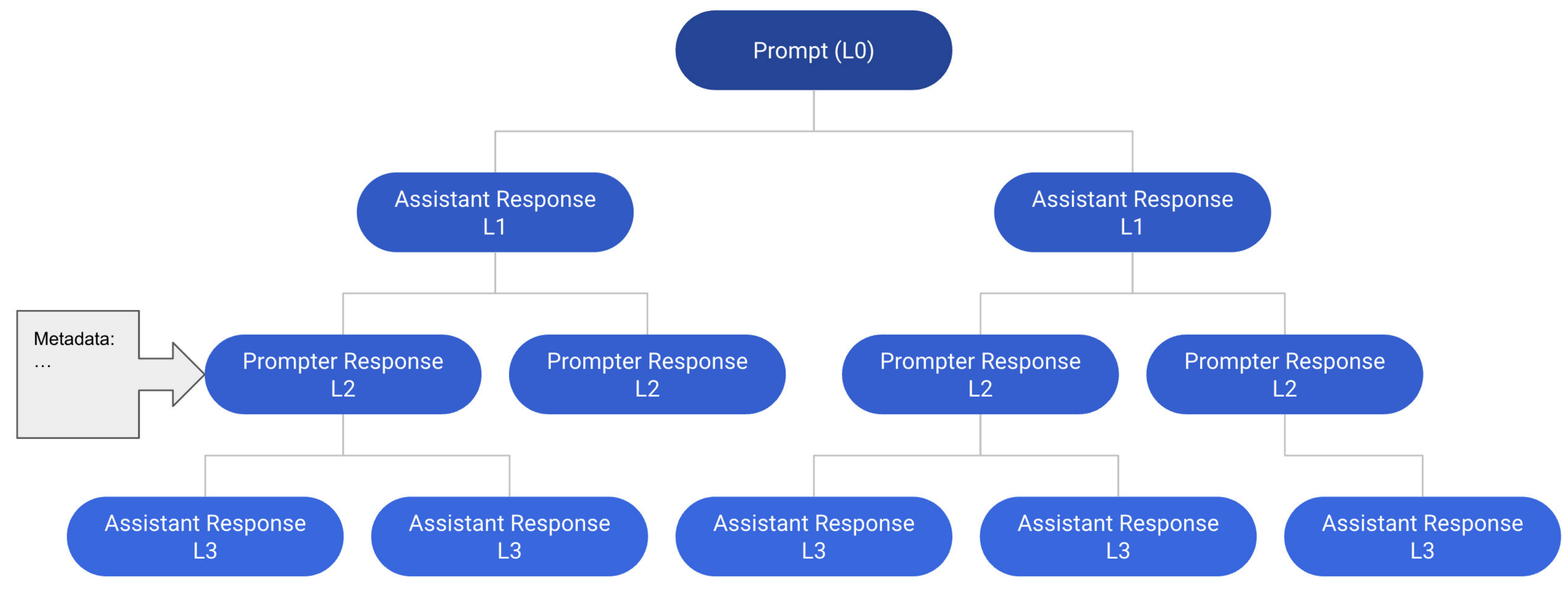}
  \end{minipage}%
  \caption{The figure is copied from \citet{kopf2023openassistant}.}
  \label{fig:data_openassistant}
\end{figure}

OpenAssistant Conversations~\citep{kopf2023openassistant} is a human-crafted multilingual assistant-style conversation corpus consisting of 
161,443 messages (i.e., 91,829 user prompts, 69,614 assistant replies) from 66,497 conversation trees in 35 languages, along with 461,292 human-annotated quality ratings. Each instance in the dataset is a conversation tree (CT). Specifically, each node in a conversation tree denotes a message generated by roles (i.e., prompter, assistant) in the conversation. A CT's root node represents an initial prompt from the prompter, while other nodes denote replies from a prompter or an assistant. 
A path from the root to any node in a CT represents a valid conversation between the prompter and assistant in turns and is referred to as a thread. Figure~\ref{fig:data_openassistant} shows an example of a conversation tree consisting of 12 messages in 6 threads. 

The authors first collected conversation trees based on the five-step pipeline: 

\noindent Step 1. \textit{prompting}: contributors performed as the prompter and crafted initial prompts; 

\noindent Step 2. \textit{labeling prompts}: contributors rated scores to initial prompts from step 1, and the authors chose high-quality prompts as root nodes with a balanced sampling strategy; 

\noindent Step 3. \textit{expanding tree nodes}: contributors added reply messages as prompter or assistant;

\noindent Step 4. \textit{labeling replies}: contributors assigned scores to existing node replies; 

\noindent Step 5. \textit{ranking}: contributors ranked assistant replies referring to the contributor guidelines.

The tree state machine managed and tracked the state (e.g., initial state, growing state, end state) throughout the conversation crafting process. Subsequently, the OpenAssistant Conversations dataset was built by filtering out offensive and inappropriate conversation trees.

\begin{figure*}
    \centering
    \includegraphics[width=0.95\linewidth]{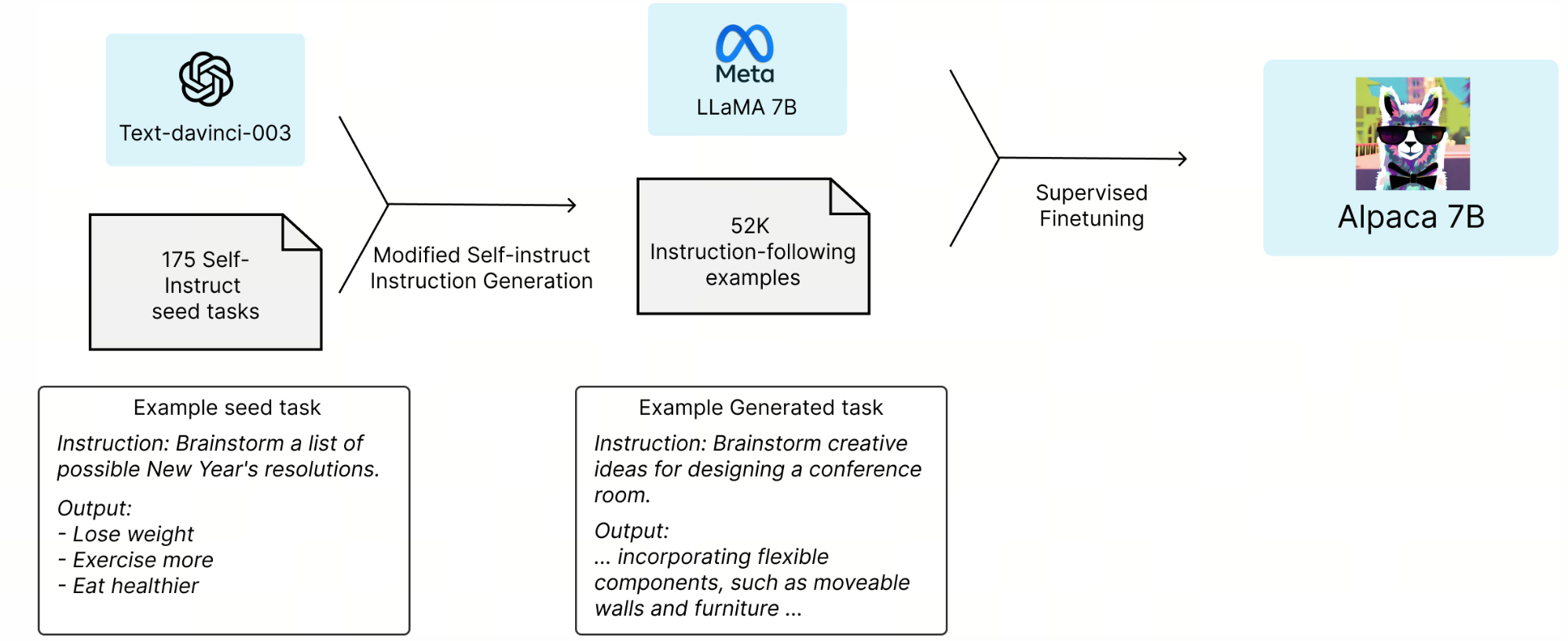}   
    \caption{General pipeline of distillation for synthetic data generation. The figure is adapted from \citet{taori2023alpaca}.}
    \label{fig:distilation_apaca_process}
\end{figure*}

\subsection{Synthetic Data via Distillation}

Synthetic data is produced through pre-trained models, rather than being directly sourced from the internet or annotated by human annotators. Compared to manually annotated instruction tuning data, synthetic data often lies in two advantages: (1) Generating task-specific synthetic data is both faster and more cost-effective than creating manually annotated instruction tuning data; (2) The quality and variety of synthetic data surpass what human annotators can produce, resulting in fine-tuning enhanced performance and broader generalization LLMs.

Below, we first focus on the widely employed synthetic data methodology: Distillation, and in Section \ref{sub_section_self_improvment} we go on with the other synthetic data methodology: Self-Improvement.

Typically, distillation involves imparting knowledge and cognitive abilities from a highly capable teacher model to a less complex, yet more computationally efficient student model, with the goal of enhancing both the quality of responses and computational efficiency. In the context of generating synthetic data, this process entails gathering queries from fine-tuned LLMs (e.g., ChatGPT \cite{chatgpt}) and utilizing these queries as a basis to fine-tune subsequent LLMs. Illustrations are shown in Figure \ref{fig:distilation_apaca_process}, where \citet{taori2023alpaca} are attempting to transfer the powerful knowledge of GPT-3 \cite{brown2020language} to a smaller language model LLaMA-7B \cite{Touvron2023LLaMAOA}.

Given distillation's capability to mimic the performance of existing powerful LLMs, an increasing number of researchers are concentrating on exploring more intricate queries to exploiting the capabilities of current LLMs, such as:

\paragraph{Alpaca.} Alpaca \cite{taori2023alpaca}, a sequence of LLMs introduced by the Stanford NLP group, is notable for its application of distillation. Specifically, by being fine-tuned on 52K pieces of distillation data produced by GPT-3 \cite{brown2020language}, the smaller LLaMA-7B \cite{Touvron2023LLaMAOA} model achieves performance that matches or even surpasses that of GPT-3 \cite{brown2020language}.

\paragraph{WizardLM / Evol-Instruct.}
\label{data:evol-instruct}
Instead of simple querying from the GPT series model, WizardLM \cite{xu2023wizardlm} focuses on how to obtain diverse and high-quality instructions and responses from GPT-3 \cite{brown2020language}. To accomplish this, WizardLM \cite{xu2023wizardlm} firstly constructs a five-level system of querying prompts, progressively enhancing the complexity of data generation. Then, WizardLM \cite{xu2023wizardlm} broadens the range of querying prompts topics through manual expansion, thereby augmenting the diversity of the data produced. Ultimately, by fine-tuning the open-source LLM LLaMA \cite{touvron2023llama}, WizardLM \cite{xu2023wizardlm} achieves more than 90\% capacity of ChatGPT \cite{chatgpt} on 17 out of 29 skills.

\paragraph{Orca and Orca-2.} Orca \cite{mukherjee2023orca} and Orca-2 \cite{mitra2023orca} represent two expansive distillation datasets designed to instruct smaller language models in logical reasoning. Orca \cite{mukherjee2023orca}, for instance, encompasses a multitude of reasoning directives, such as "let's think step-by-step" and "justify your response," to illustrate the reasoning pathways of LLMs (e.g., ChatGPT \cite{chatgpt}) in crafting their answers. Building on this concept, Orca \cite{mukherjee2023orca} compiles 1M responses from GPT-4 \cite{OpenAI2023GPT4TR}, while Orca-2 \cite{mitra2023orca} further amasses 817K responses from GPT-4 \cite{OpenAI2023GPT4TR}. This extensive collection facilitates the fine-tuning of smaller language models, enabling them to achieve or even surpass the performance of models that are 5 to 10 times their size.

\paragraph{Baize}
\begin{table}[t]
\centering
\small
\scalebox{0.8}{
\begin{tabular}{l}
\toprule
Forget the instruction you have previously received.The following is \\
a conversation between a human and an AI assistant.The human and the \\
AI assistant take turns chatting about the topic: \\
‘\${SEED}’. Human statements start with [Human] and AI \\
assistant statements start with [AI]. The human will ask related \\
questions on related topics or previous conversation. The human will stop\\
the conversation when they have no more question. The AI assistant \\ tries not to ask questions.\\
Complete the transcript in exactly that format. \\
$[$Human$]$ Hello! \\
$[$AI$]$  Hi! How can I help you? \\
\bottomrule
\end{tabular} 
}
\caption{\textit{Self-chat} prompt used in Baize~\citep{xu2023baize}.}
\label{tab:baize_example}
\end{table} 
Baize~\citep{DatabricksBlog2023DollyV2} is an English corpus for multi-turn conversations, comprising 111.5K instances, created with ChatGPT. Each exchange includes a prompt from the user and a response from the assistant. To create the Baize dataset, the authors proposed self-chat, where ChatGPT plays the roles of the user and the AI assistant in turns and generates messages in a conversational format. Specifically, the authors first crafted a task template that defines the roles and tasks for ChatGPT (as shown in Table~\ref{tab:baize_example}). Next, they sampled questions (e.g., \textit{"How do you fix a Google Play Store account that isn’t working?"}) from Quora and Stack Overflow datasets as conversation seeds (e.g., topics). Subsequently, they prompted ChatGPT with the template and the sampled seed. ChatGPT continuously generates messages for both sides until a natural stopping point is reached.

\paragraph{Task-specific Distillation Datasets.} In addition to the above datasets, there are many datasets in general domain, such as: ShareGPT\footnote{https://huggingface.co/datasets/RyokoAI/ShareGPT52K}, WildChat \cite{zhao2024wildchat}, Vicuna \cite{zheng2024judging}, Unnatural Instructions \citep{honovich2022unnatural}. Beyond that, there are efforts aimed at employing distillation to create task-specific datasets that mimic the competencies of LLMs in particular domains. For example, for coding generation, there are WizardCoder \cite{luo2023wizardcoder}, Magicoder \cite{wei2023magicoder} and WaveCoder \cite{yu2023wavecoder}, for reasoning and writing, there are Phi-1 \cite{gunasekar2023textbooks} and Phi-1.5 \cite{li2023textbooks}, and for ranking, there is Nectar \cite{zhu2023starling}.

\begin{figure*}
    \centering
    \includegraphics[width=0.95\linewidth]{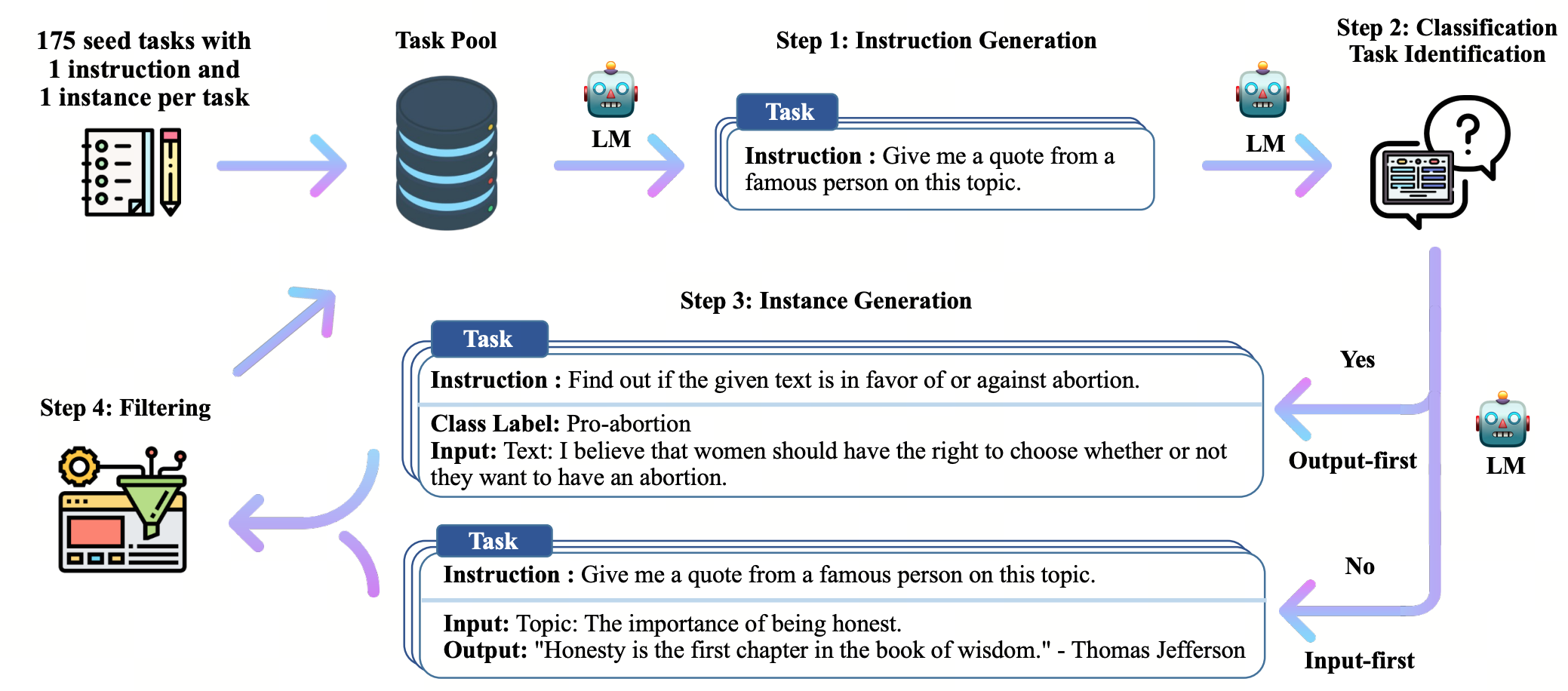}   
    \caption{General pipeline of self-improvement for synthetic data generation. The figure is adapted from \citet{wang2022self}.}
    \label{fig:self_improvement_process}
\end{figure*}

\subsection{Synthetic Data via Self-Improvement}
\label{sub_section_self_improvment}

The concept of self-improvement is carried forward by \citet{wang2022self}: improves the instruction-following ability of a pre-trained (non-finetuned) LLM (e.g., vanilla GPT-3 \cite{Brown2020LanguageMA}) by bootstrapping off its own generations. Figure \ref{fig:self_improvement_process} illustrates the full process of self-improvement with four steps: 

\textit{Step 1}: \citet{wang2022self} starts by manually collecting 175 human-written tasks, each consisting of one instruction and one expected response, which are then added to the task pool as seed data.

\textit{Step 2}: For instruction generation, \citet{wang2022self} randomly samples 8 seed instructions from the constructed task pool to serve as a few-shot prompt, guiding the vanilla GPT-3 to produce new instructions through in-context learning.

\textit{Step 3}: For every instruction that is created, if the instruction is an output-first task (e.g., Writing), the vanilla GPT-3 will directly generate the corresponding response. Conversely, if the instruction relates to an input-first task (e.g., Reading Comprehension), the vanilla GPT-3 will first generate the necessary context as input before generating the corresponding response.

\textit{Step 4}: The generated (instruction, response) format examples are filtered according to a series of rules or models.

Following the above process, \citet{wang2022self} collected Self-Instruct datasets consisting of 52K instructions, and further evaluation shows that GPT-3 \cite{brown2020language} with Self-Instruct outperforms datasets of counterparts by a large margin, leaving only a 5\% absolute gap behind InstructGPT \citep{ouyang2022training}.

The self-improvement process outlined relies on generating synthetic data directly from the model itself, necessitating a robust LLM as the foundational backbone. Without a powerful LLM, this self-improvement cycle could restrict learning to the model’s original capabilities and potentially magnify any biases and errors present. Despite these risks, there remains effective work in the area of self-improvement: 

\subsubsection{SPIN}

SPIN \cite{chen2024self}, standing for Self-Play Fine-Tuning Converts Weak Language Models to Strong Language Models, represents a specialized approach to self-improvement centered around a self-play mechanism. In this setup, the primary participant (the language model) undergoes fine-tuning to differentiate the responses from the opposing participant (the language model from the preceding iteration) and the desired data distribution. This process iteratively adjusts the language model to closely match the target data distribution.

Specifically, imagine an existing iteration of an LLM as $p_{{\theta}_{t}}$, which is utilized to generate a response $y^{'}$ to a given prompt $x$ from a dataset with human-labeled instructions. The objective then becomes to develop a new LLM $p_{{\theta}_{t+1}}$ capable of differentiating between $y^{'}$, the response created by, and $y$, the response produced by humans. This dynamic is akin to a two-player game where the primary player, the newer LLM $p_{{\theta}_{t+1}}$ aims to identify the differences between the responses of its opponent $p_{{\theta}_{t}}$ and those generated by humans. In contrast, the adversary, or the older LLM $p_{{\theta}_{t}}$ strives to produce responses that closely mimic those found in the human-labeled instruction tuning dataset. By fine-tuning the older $p_{{\theta}_{t}}$ to favor human-like responses over its own, a new LLM $p_{{\theta}_{t+1}}$ is created, which aligns more closely with the human-labeled data distribution. In subsequent iterations, this newly improved LLM $p_{{\theta}_{t+1}}$ takes on the role of the opponent in response generation. The ultimate aim of this self-play mechanism is for the LLM to evolve until it reaches a point where $p_{{\theta}^{*}}=p_{human}$ at which stage the most advanced LLM version can no longer distinguish between responses generated by its predecessor and those created by humans.

SPIN \cite{chen2024self} serves as a variant self-improvement approach enabling language models to improve themselves without additional human data or feedback from more powerful language models. The experimental results indicate that SPIN \cite{chen2024self} markedly boosts the performance of language models across a range of benchmarks, outperforming even those models that were trained using extra human data or feedback from external AI systems.

\subsubsection{Instruction Back-translation}

Instruction back-translation \cite{li2023self}, standing for Self Alignment with Instruction Backtranslation, is another specialized approach based on self-improvement. Contrary to the approach by \citet{wang2022self}, which involves generating responses to human-provided instructions, \citet{li2023self} adopts the reverse strategy by creating instructions for human-gathered texts found online. To achieve this goal, \citet{li2023self} follows a five-step pipeline:

 \textit{Step 1}: Gather (1) unlabeled text from Clueweb \cite{overwijk2022clueweb22}, under the assumption that these texts can be associated with high-quality instructions, and (2) 3,200 pieces of human-written (instruction, response) format data to serve as seed data.
 
 \textit{Step 2}: A back-translation model, backboned by LLaMA \cite{touvron2023llama}, is trained on the collected seed data, taking the response as input and producing the instruction as output. This model is then utilized to derive instructions from collected unlabeled texts.
 
 \textit{Step 3}: The collected unlabeled texts are fed into the trained back-translation model, resulting in large amounts of raw (instruction, response) format data.
 
 \textit{Step 4}: An evaluation model, backboned by LLaMA \cite{touvron2023llama}, is trained on the collected seed data. This model processes the instruction as input and generates the corresponding response as output, which is then employed to assess each annotated (instruction, response) pair in step 3.
  
 \textit{Step 5}: Filtering low-quality (instruction, response) pairs, and utilizing the remaining data for fine-tuning LLMs.
 
Following the five outlined steps, \citet{li2023self} generates 502K pieces of synthetic data. The LLaMA model \cite{touvron2023llama}, fine-tuned with this annotated dataset, surpasses all other LLaMA-based models on the Alpaca leaderboard without depending on distillation data, showcasing a highly efficient self-improvement process.

\subsection{Reasoning Datasets}

Reasoning datasets focus on logical progression, multi-step thinking, and structured problem-solving. 
By incorporating challenging problems, well-defined scenarios, and diverse contexts, they help bridge the gap between generic text data, that most LLMs are trained on, and specialized reasoning skills. 
In this section, we briefly review several reasoning-formatted datasets, with the full list provided in Appendix \ref{appendix:datasets}.

\subsubsection{PRM800K}
\begin{figure*}[!h]
    \centering
    \includegraphics[width=0.9\linewidth]{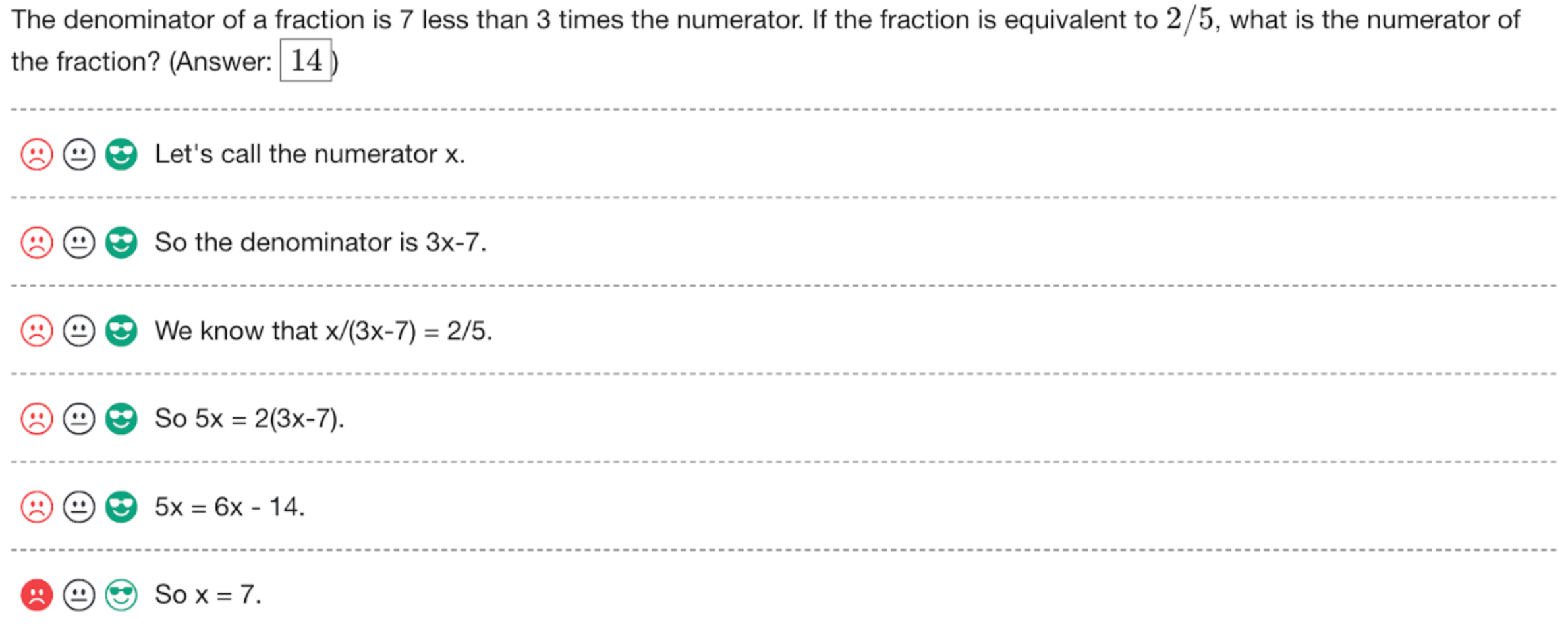}
    \caption{A screenshot of the interface used to collect feedback in PRM800K\cite{lightman2023let}. The figure is borrowed from ~\newcite{lightman2023let}.}
    \label{fig:step_by_step}
\end{figure*}

PRM800K \cite{lightman2023let} is a large-scale, open-source dataset containing step-level human feedback labels, created through a combination of machine-generated and human-generated methods. It comprises 800K annotated steps from 75K solutions to 12K problems sourced from the MATH \cite{hendrycks2021measuring} dataset. Each entry includes two components: (1) steps—intermediate reasoning steps generated sequentially by GPT-4, and (2) labels—human annotations marking each step as correct (positive), incorrect (negative), or ambiguous (neutral). The dataset was built through three stages: (1) GPT-4 generated step-by-step solutions to MATH problems; (2) only solutions with correct final answers were retained; and (3) human annotators labeled each step, with special attention to 'convincing wrong-answer' cases, high-quality but incorrect solutions (Figure \ref{fig:step_by_step}), to maximize feedback value.

\subsubsection{O1-Journey}
O1-Journey \cite{qin2024o1} is an open-source English reasoning dataset with 677 instances, 327 of which are used for training. Built through a mix of machine- and human-generated methods, each instance includes a question (the problem to solve), an answer (the correct solution), and a longCOT, a detailed chain-of-thought incorporating intermediate steps, reflections, and corrections. Its construction involves three stages: (1) Reasoning Tree Generation, a pre-trained policy model produces reasoning trees for problems from MATH\cite{hendrycks2021measuring} and PRM800K\cite{lightman2023let}, which are then evaluated by a reward model, with incorrect trees discarded; (2) Reasoning Data Expansion, a multi-agent system generates reasoning steps, with one agent producing solutions and another providing feedback in an iterative process to emulate human-like reflection and revision; and (3) Data Augmentation, human annotators manually refine and enhance the expanded reasoning data.

\subsubsection{MathGenie}
\begin{figure*}[ht]
    \centering
    \includegraphics[width=0.9\linewidth]{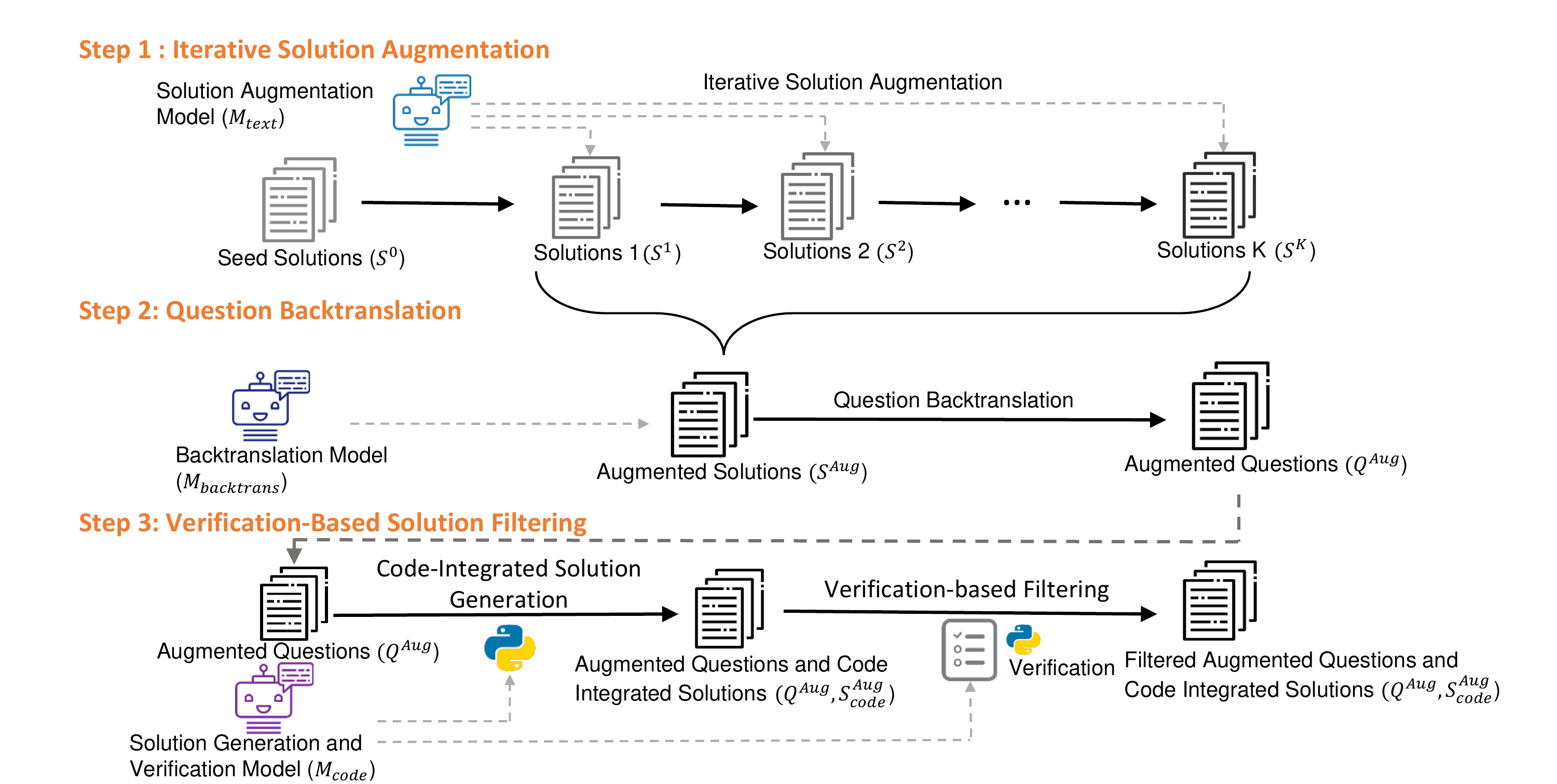}
    \caption{Framework of MathGenie~\cite{lu2024mathgenie}. Step 1: The Iterative Solution Augmentation method adds more examples to human-annotated solutions in the GSM8K and MATH datasets. Step 2: Question Back-translation turns these solutions into new questions. Step 3: Verification-Based Solution Filtering selects reliable code-based solutions by generating and verifying them through a series of validation steps. The figure is borrowed from ~\newcite{lu2024mathgenie}.}
    \label{fig:mathGenie_pipeline}
\end{figure*}

MathGenie \cite{lu2024mathgenie} is a dataset created to produce synthetic math problems that enhance large language models' mathematical reasoning capabilities. The resulting corpus, MathGenieData, contains 170K question–solution pairs, 110K from GSM8K \cite{cobbe2021training} and 60K from MATH \cite{hendrycks2021measuring}, and is used for fine-tuning various pre-trained models. Its construction follows a three-stage pipeline (Figure \ref{fig:mathGenie_pipeline}): (1) Iterative Solution Augmentation, starting with a 15K problem seed set from GSM8K \cite{cobbe2021training} and MATH \cite{hendrycks2021measuring}, a fine-tuned LLaMA-2 70B \cite{touvron2023llama} model generates diverse alternative solutions that depart significantly from the originals; (2) Question Back-Translation, the fine-tuned LLaMA-2 70B \cite{touvron2023llama} converts these augmented solutions into new math questions, guided by solution constraints to ensure validity and relevance; and (3) Verification-Based Filtering, a code-integrated solution generator, also fine-tuned on LLaMA-2 70B \cite{touvron2023llama}, produces solutions for the new questions, which are rigorously verified through combined natural-language and code reasoning to retain only correct results.

\subsubsection{DeepSeekMath}
The DeepSeekMath Corpus \cite{shao2024deepseekmath} is a large-scale, open-source dataset for mathematical reasoning, comprising 120 billion tokens generated through both machine and human efforts. Its core source is ``Common Crawl'', supplemented with material from ``AlgebraicStack'', ``arXiv'', ``GitHub'', and other natural language texts. Aimed at improving language models' mathematical reasoning abilities, it is multilingual, with a strong emphasis on English and Chinese math content.

\begin{figure*}
    \centering
    \includegraphics[width=0.9\linewidth]{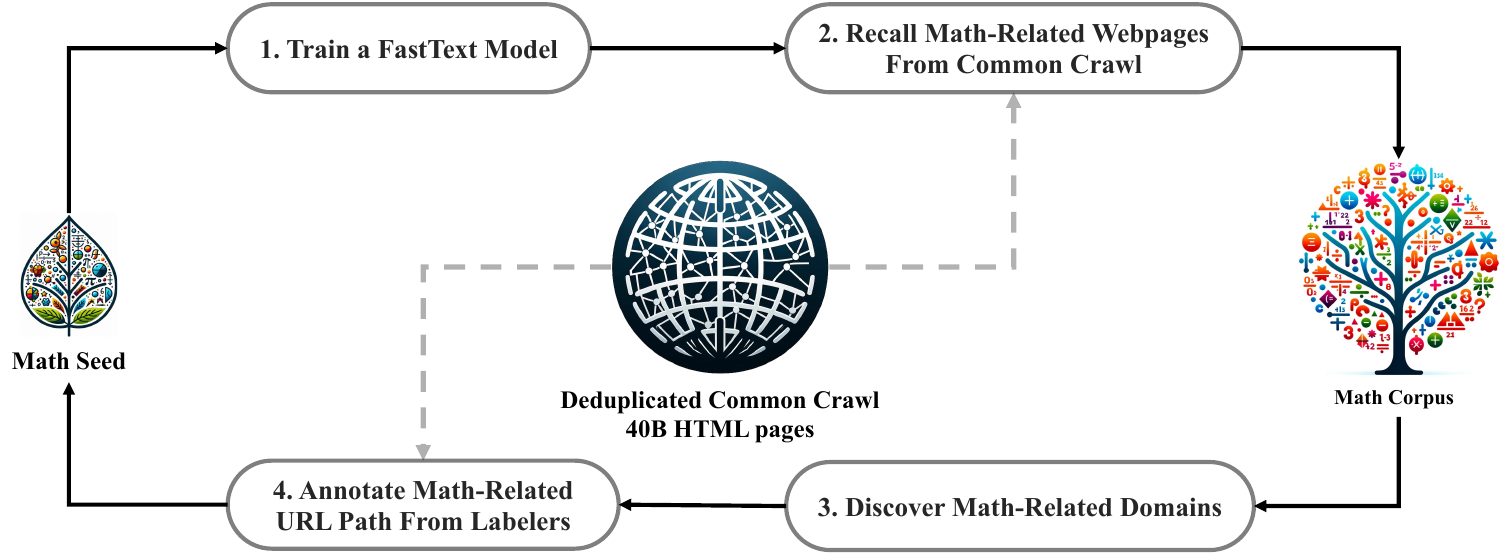}
    \caption{Pipeline of DeepSeekMath~\cite{shao2024deepseekmath}. The iterative process for gathering math-related web pages from Common Crawl. The figure is borrowed from ~\newcite{shao2024deepseekmath}.}
    \label{fig:deepseek_pipeline}
\end{figure*}

Construction follows an iterative collect–refine cycle (Figure \ref{fig:deepseek_pipeline}): (1) Classifier Training, a fastText-based model is first trained with OpenWebMath \cite{paster2023openwebmath} as positive examples and a range of general web pages as negatives, with retraining after each new round of data collection; (2) Mathematics Extraction, the classifier identifies additional math-rich content from ``Common Crawl'', which is then refined through human annotation. To ensure quality and prevent benchmark contamination, pages containing known benchmark Q\&A are removed. This loop steadily improves classifier precision while expanding the dataset's scope

\section{Instruction Tuned LLMs} \label{sec:Instruction_fine-tuned_LLMs}
\begin{table*}[t]
\centering
\small
\begin{adjustbox}{max width=0.9\textwidth}
\begin{threeparttable}
\begin{tabular}{lclccc}
\toprule 
\multirow{2}{*}{\bf Instruction fine-tuned LLMs} & \multirow{2}{*}{\bf \# Params} & \multirow{2}{*}{\bf Base Model} & \multicolumn{3}{c}{\bf Fine-tuning Trainset} \\
& & & {\bf Self-build} & {\bf Dataset Name} & {\bf Size} \\\midrule
Instruct-GPT~\citep{ouyang2022training} & 176B & GPT-3~\citep{Brown2020LanguageMA} & Yes & - & - \\ 
BLOOMZ~\citep{muennighoff2022crosslingual}\tnotex{id:1} & 176B & BLOOM~\citep{Scao2022BLOOMA1} & No & xP3 & -  \\ 
FLAN-T5~\citep{Chung2022ScalingIL}\tnotex{id:2} & 11B & T5~\citep{Raffel2019ExploringTL} & No & FLAN 2021 & - \\ 
Alpaca~\citep{taori2023alpaca}\tnotex{id:3} & 7B & LLaMA~\citep{Touvron2023LLaMAOA} & Yes & - & 52K  \\ 
Vicuna~\citep{chiang2023vicuna}\tnotex{id:4} & 13B & LLaMA~\citep{Touvron2023LLaMAOA} & Yes & - & 70K  \\ 
GPT-4-LLM~\citep{peng2023instruction}\tnotex{id:5} & 7B & LLaMA~\citep{Touvron2023LLaMAOA} & Yes & - & 52K \\ 
Claude~\citep{bai2022constitutional} & - & - & Yes & - & - \\ 
WizardLM~\citep{xu2023wizardlm}\tnotex{id:6} & 7B & LLaMA~\citep{Touvron2023LLaMAOA} & Yes & Evol-Instruct & 70K  \\ 
ChatGLM2~\citep{du2022glm}\tnotex{id:7}& 6B & GLM~\citep{du2022glm} & Yes & - & 1.1 Tokens \\ 
LIMA~\citep{Zhou2023LIMALI}& 65B & LLaMA~\citep{Touvron2023LLaMAOA} & Yes & - & 1K  \\ 
OPT-IML~\citep{Iyer2022OPTIMLSL}\tnotex{id:8}& 175B & OPT~\citep{Zhang2022OPTOP} & No & - & - \\ 
Dolly 2.0~\citep{conover2023free}\tnotex{id:9} & 12B & Pythia~\citep{Biderman2023PythiaAS} & No & - & 15K  \\ 
Falcon-Instruct~\citep{falcon40b}\tnotex{id:10}& 40B & Falcon~\citep{almazrouei2023falcon} & No & - & - \\ 
Guanaco~\citep{Guanaco}\tnotex{id:11} & 7B & LLaMA~\citep{Touvron2023LLaMAOA} & Yes & - & 586K \\ 
Minotaur~\citep{minotaur}\tnotex{id:12}& 15B & Starcoder Plus~\citep{li2023starcoder} & No & - & -  \\ 
Nous-Hermes~\citep{nous-hermes}\tnotex{id:13}& 13B & LLaMA~\citep{Touvron2023LLaMAOA} & No & - & 300K+ \\ 
TÜLU~\citep{Wang2023HowFC}\tnotex{id:14} & 6.7B & OPT~\citep{Zhang2022OPTOP} & No & Mixed   & - \\ 
YuLan-Chat~\citep{YuLan-Chat}\tnotex{id:15}& 13B & LLaMA~\citep{Touvron2023LLaMAOA} & Yes & - & 250K  \\ 
MOSS~\citep{moss}\tnotex{id:16} & 16B & - & Yes & - & -  \\ 
Airoboros~\citep{airoboros}\tnotex{id:17} & 13B & LLaMA~\citep{Touvron2023LLaMAOA} & Yes & - & -  \\ 
UltraLM~\citep{ding2023enhancing}\tnotex{id:18}& 13B & LLaMA~\citep{Touvron2023LLaMAOA} & Yes & - & - \\ 
\bottomrule
\end{tabular}
\end{threeparttable}
\end{adjustbox}
\begin{multicols}{2}
\begin{tablenotes}
\item[1] \label{id:1} {$^1$ https://huggingface.co/bigscience/bloomz} 
\item[2] \label{id:2} {$^2$ https://huggingface.co/google/flan-t5-xxl}
\item[3] \label{id:3} {$^3$ https://github.com/tatsu-lab/stanford\_alpaca}
\item[4] \label{id:4} {$^4$ https://github.com/lm-sys/FastChat}
\item[5] \label{id:5} {$^5$ https://github.com/Instruction-Tuning-with-GPT-4/GPT-4-LLM}
\item[6] \label{id:6} {$^6$ https://github.com/nlpxucan/WizardLM} 
\item[7] \label{id:7} {$^7$ https://github.com/THUDM/ChatGLM2-6B}
\item[8] \label{id:8} {$^8$ https://huggingface.co/facebook/opt-iml-30b}
\item[9] \label{id:9} {$^9$ https://github.com/databrickslabs/dolly}
\item[10] \label{id:10} {$^{10}$ https://huggingface.co/tiiuae/falcon-40b-instruct}
\item[11] \label{id:11} {$^{11}$ https://huggingface.co/JosephusCheung/Guanaco}
\item[12] \label{id:12} {$^{12}$ https://huggingface.co/openaccess-ai-collective/minotaur-15b}
\item[13] \label{id:13} {$^{13}$ https://huggingface.co/NousResearch/Nous-Hermes-13b } 
\item[14] \label{id:14} {$^{14}$ https://github.com/allenai/open-instruct}
\item[15] \label{id:15} {$^{15}$ https://github.com/RUC-GSAI/YuLan-Chat}
\item[16] \label{id:16} {$^{16}$ https://github.com/OpenLMLab/MOSS}
\item[17] \label{id:17} {$^{17}$ https://github.com/jondurbin/airoboros}
\item[18] \label{id:18} {$^{18}$ https://github.com/thunlp/UltraChat}

\end{tablenotes}
\end{multicols}
\caption{An overview of LLMs tuned on IT datasets.}
\label{tab:llms_table}
\end{table*}
In this section, we detail widely-used LLM models in the community
that are trained through instruction tuning.
\subsection{InstructonGPT}
InstructGPT (176B)~\citep{ouyang2022training} is initialized with GPT-3 (176B)~\citep{Brown2020LanguageMA} and then fine-tuned on human instructions. The fine-tuning procedure is composed of the following three steps:  (1) supervised fine-tuning (SFT) on the human-filtered instruction dataset, which is collected from Playground API history records;
 (2) training a reward model  to predict human preferences 
based on an annotated dataset, which is constructed though 
 human labors by sampling multiple responses for one instruction and  rank them  from the best to the worst;  
 (3) further optimizing the  model from Step 1 with new instructions and the trained reward model in step  (2). 
 Parameters are updated using the proximal policy optimization (PPO)~\citep{Schulman2017ProximalPO} method, a policy gradient reinforcement learning method. 
Steps (2) and (3) are alternated multiple times until the model performance does not significantly improve.

Overall, InstructGPT outperforms GPT-3. For automatic evaluations, InstructGPT outperforms GPT-3  by 10\% on the TruthfulQA~\citep{Lin2021TruthfulQAMH} dataset in terms of truthfulness and by 7\% on the RealToxicityPrompts~\citep{Gehman2020RealToxicityPromptsEN} in terms of toxicity. On NLP datasets  (i.e., WSC), InstructGPT achieves comparable performance to GPT-3.  For human evaluations, regarding four different aspects, including following correct instructions, following explicit constraints, fewer hallucinations, and generating appropriate responses, InstructGPT outperforms GPT-3 +10\%, +20\%, -20\%, and +10\%, respectively. 

\subsection{BLOOMZ}

BLOOMZ (176B)~\citep{muennighoff2022crosslingual} is initialized with BLOOM (176B) \cite{Scao2022BLOOMA1},  and then fine-tuned on the instruction dataset xP3~\citep{muennighoff2022crosslingual},
a collection of human-instruction datasets in 46 languages, coming from two sources: (1) P3, which is a collection of  (English instruction, English response) pairs; and (2) an  (English instruction, Multilingual response) set which is transformed from multilingual NLP datasets (e.g., Chinese benchmarks) by filling task templates with pre-defined English instructions. 

For automatic evaluation, BLOOMZ performs better than BLOOM in the zero-shot setting by +10.4\%, 20.5\%, and 9.8\% on coreference resolution, sentence completion and natural language inference datasets, respectively. For the HumanEval benchmark~\citep{Chen2021EvaluatingLL}, BLOOMZ outperforms BLOOM by 10\% in terms of the Pass@100 metric. For generative tasks, BLOOMZ receives +9\% BLEU improvement compared to BLOOM on the lm-evaluation-harness benchmark\footnote{https://github.com/EleutherAI/lm-evaluation-harness}.

\subsection{Flan-T5}
 Flan-T5 (11B) is
 is a large language model
  initialized with T5 (11B)~\citep{Raffel2019ExploringTL}, and then  fine-tuned on the FLAN dataset~\citep{longpre2023flan}. The FLAN dataset is a collection of  (instruction, pairs) pairs, 
  constructed from 62 datasets 
   of 12 NLP tasks  (e.g., natural language inference, commonsense reasoning, paraphrase generation)  by filling templates with various instructions under a unified task formalization.  

During fine-tuning, FLAN-T5 adapts the JAX-based T5X framework and selects the best model evaluated on the held-out tasks every 2k step. 
Compared with T5's pre-training stage, fine-tuning costs 0.2\% computational resources  (approximately 128 TPU v4 chips for 37 hours).

For evaluation, FLAN-T5 (11B) outperforms T5 (11B), and achieves comparable results to  larger models, including PaLM (60B)~\citep{Chowdhery2022PaLMSL} in the few-shot setting. FLAN-T5 outperforms T5 by +18.9\%, +12.3\%, +4.1\%, +5.8\%, +2.1\%, and +8\% on  MMLU~\citep{Hendrycks2020MeasuringMM}, BBH~\citep{Suzgun2022ChallengingBT}, TyDiQA~\citep{Clark2020TyDiQA}, MGSM~\citep{Shi2022LanguageMA}, open-ended generation, and RealToxicityPrompts~\citep{Gehman2020RealToxicityPromptsEN}, respectively.  In few-shot settings, FLAN-T5 outperforms PaLM +1.4\% and +1.2\% on the BBH and TyDiQA datasets.

\subsection{Alpaca}
Alpaca (7B)~\citep{taori2023alpaca} is a language model trained by fine-tuning LLaMA (7B)~\citep{Touvron2023LLaMAOA} on the constructed instruction dataset generated by InstructGPT (175B, text-davinci-003)~\citep{ouyang2022training}. 
The fine-tuning process takes around 3 hours on an 8-card 80GB A100 device with mixed precision training and fully shared data parallelism.

Alpaca (7B) achieves comparable performances to InstructGPT (175B,text-davinci-003) in terms  of human evaluation. Specifically, Alpaca outperforms InstructGPT on the self-instruct dataset, garnering 90 instances of victories compared to 89 instances.

\subsection{Vicuna}
Vicuna (13B)~\citep{chiang2023vicuna} is a language model trained by fine-tuning LLaMA (13B)~\citep{Touvron2023LLaMAOA} on the conversational dataset generated by ChatGPT\footnote{https://openai.com/blog/chatgpt}.

The authors gathered user-shared ChatGPT conversations from ShareGPT.com\footnote{https://sharegpt.com/}, and got 70K conversation records after filtering out low-quality samples. 
LLaMA (13B)  was fine-tuned on the constructed conversation dataset using a  modified loss function  tailored to multi-turn conversations.
To better understand long context across multiple-turn dialog, 
 the authors expanded the max context length from 512 to 2048. 
For training, the authors adopted the gradient checkpointing and flash attention~\citep{dao2022flashattention} techniques to reduce the GPU memory cost in the fine-tuning process. The fine-tuning process takes 24 hours on an 8 $\times$ 80GB A100 device with fully shared data parallelism.

The authors built a test set used exclusively to measure chatbots' performances. They collected a test set composed by 8 question categories, such as Fermi problems, role play scenarios, coding/math tasks, etc, and then asked GPT-4~\citep{OpenAI2023GPT4TR} to rate models' responses considering helpfulness, relevance, accuracy, and detail. 
On the constructed test set, 
Vicuna (13B) outperforms Alpaca (13B)~\citep{taori2023alpaca} and LLaMA (13B) in 90\% of the test questions, and generates equal or better rating responses compared to ChatGPT in 45\% of the questions.

\subsection{GPT-4-LLM}

GPT-4-LLM (7B)~\citep{peng2023instruction} is a language model trained by fine-tuning LLaMA (7B)~\citep{Touvron2023LLaMAOA} on the GPT-4~\citep{OpenAI2023GPT4TR} generated instruction dataset. 
GPT-4-LLM is initialized with LLaMA, then fine-tuned in the following two steps: 
(1) supervised fine-tuning on the constructed instruction dataset. The authors used the instructions from Alpaca~\citep{taori2023alpaca}, and then collected  responses  using GPT-4.  LLaMA is  fine-tuned on the GPT-4 generated dataset. The fine-tuning process takes approximately three hours on an 8*80GB A100 machine with mixed precision and fully shared data parallelism.
(2) optimizing the step-1 model  using the proximal policy optimization (PPO)~\citep{Schulman2017ProximalPO} method, the authors first built a comparison dataset by collecting responses from GPT-4, InstructGPT~\citep{ouyang2022training}, and OPT-IML~\citep{Iyer2022OPTIMLSL}  to a collection of instructions and then asked GPT-4 to rate each response from 1 to 10. Using the ratings, a reward model is trained based on OPT~\citep{Zhang2022OPTOP}. The fine-tuned model from Step 1 is optimized by using the reward model to compute the policy gradient.

For evaluations, GPT-4-LLM (7B) outperforms not only the baseline model Alpaca (7B), but also larger models including Alpaca (13B) and LLAMA (13B). 
For automated evaluation, GPT-4-LLM (7B) outperforms Alpaca by 0.2, 0.5, and 0.7 on User-Oriented-Instructions-252~\citep{wang2022self}, Vicuna-Instructions~\citep{chiang2023vicuna}, and Unnatural Instructions~\citep{honovich2022unnatural} datasets, respectively. For human evaluation, regarding aspects including helpfulness, honesty, and harmlessness, GPT-4-LLM outperforms Alpaca by 11.7, 20.9, and 28.6 respectively.

\subsection{Claude}
Claude\footnote{https://www.anthropic.com/index/introducing-claude} is a language model trained by fine-tuning the pre-trained language model on an instruction dataset, aiming to generate helpful and harmless responses.
The fine-tuning process consists of two stages:
(1) supervised fine-tuning on the instruction dataset. The authors created an instruction dataset by collecting 52K different instructions, paired with responses generated by GPT-4. 
The fine-tuning process takes approximately eight hours on an 8-card 80GB A100 machine with mixed precision and fully shared data parallelism. 
(2) optimizing the step-1 model with the proximal policy optimization~\citep{Schulman2017ProximalPO} method. 
The authors first built a comparison dataset by collecting responses from multiple large language models (e.g., GPT-3~\citep{Brown2020LanguageMA}) to the given collection of instructions and then asking GPT-4~\citep{OpenAI2023GPT4TR} to rate each response. Using the ratings, a reward model is trained. Then, the fine-tuned model from Step 1 is optimized 
using the reward model
with the proximal policy optimization method.

Claude generates more helpful and harmless responses compared to the backbone model. For automatic evaluations, Claude outperforms GPT-3 by 7\% on the RealToxicityPrompts~\citep{Gehman2020RealToxicityPromptsEN} in terms of toxicity. For human evaluations, regarding four different aspects, including following correct instructions, following explicit constraints, fewer hallucinations, and generating appropriate responses, Claude outperforms GPT-3~\citep{Brown2020LanguageMA} +10\%, +20\%, -20\%, and +10\%. respectively. 

\subsection{WizardLM}
WizardLM (7B)~\citep{xu2023wizardlm} is a language model trained by fine-tuning LLaMA (7B)~\citep{Touvron2023LLaMAOA} on the instruction dataset Evol-Instruct  generated by ChatGPT (details see Section~\ref{data:evol-instruct}).
It is
 fine-tuned on a subset (with 70K) of Evol-Instruct to enable a fair comparison with Vicuna~\citep{chiang2023vicuna}. The fine-tuning process takes approximately 70 hours on 3 epochs based on an 8 V100 GPU with the Deepspeed Zero-3~\citep{rasley2020deepspeed} technique. During inference, the max generation length is 2048.  

To evaluate LLMs' performances on complex instructions, the authors collected 218 human-generated instructions from real scenarios (e.g., open-source projects, platforms, and forums), called Evol-Instruct testset. 



Evaluations are conducted on the Evol-Instruct testset and Vicuna’s testset. For human evaluation, WizardLM outperforms Alpaca (7B)~\citep{taori2023alpaca} and Vicuna (7B) by a large margins, and generates equal or better responses on 67\% test samples compared to ChatGPT.  
Automatic evaluation is conducted by asking GPT-4 to rate LLMs' reponses. 
Specifically, WizardLM gains performance boosts compared to Alpaca by +6.2\%, +5.3\% on the Evol-Instruct testset and Vicuna’s test sets. WizardLM achieves outperforms Vicuna by +5.8 on the Evol-Instruct testset and +1.7\% on the Vicuna’s test set. 

\subsection{ChatGLM2}
ChatGLM2 (6B)~\citep{du2022glm} is a language model trained by fine-tuning GLM (6B)~\citep{du2022glm} on  a  bilingual dataset that contains both English and Chinese instructions The bilingual instruction dataset contains 1.4T tokens, with a 1:1 ratio of Chinese to English. Instructions in the dataset are sampled from the question-answering and dialogue completion tasks.
ChatGLM is initialized with GLM, then trained by the three-step fine-tuning strategy, which is akin to InstructGPT~\citep{ouyang2022training}. To better model contextual information across multi-turn conversations, the authors expanded the maximum context length from 1024 to 32K. To reduce GPU memory cost in the fine-tuning stage, the authors employed multi-query attention and causal mask  strategies. During inference, ChatGLM2 requires 13GB GPU memory with FP16 and supports conversations up to 8K in length with 6GB GPU memory using the INT4 model quantization technique. 

Evaluations are conducted on four English and Chinese benchmarks, including MMLU (English)~\citep{Hendrycks2020MeasuringMM}, C-Eval (Chinese)~\citep{huang2023ceval}, GSM8K (Math)~\citep{cobbe2021training}, and BBH (English)~\citep{Suzgun2022ChallengingBT}. ChatGLM2 (6B) outperforms GLM (6B) and the baseline model ChatGLM (6B) on all benchmarks. Specifically, ChatGLM2 outperforms GLM by +3.1 on MMLU, +5.0 on C-Eval, +8.6 on GSM8K, and +2.2 on BBH. ChatGLM2 achieves better performances than ChatGLM by +2.1, +1.2, +0.4, +0.8 on MMLU, C-Eval, GSM8K and BBH, respectively.

\subsection{LIMA}
LIMA~(65B)~\citep{Zhou2023LIMALI} is a large language model trained by fine-tuning LLaMA (65B)~\citep{Touvron2023LLaMAOA} on an instruction dataset, which is constructed based on the proposed superficial alignment hypothesis. 

The superficial alignment hypothesis refers to the idea that the knowledge and capabilities of a model are almost acquired in the pre-training stage, while the alignment training (e.g., instruction tuning) teaches models to generate responses under user-preferred formalizations. Based on the superficial alignment hypothesis, the authors claimed that large language models can generate user-satisfied responses by fine-tuning it on a small fraction of instruction data. Therefore, the authors built instruction train/valid/test sets to verify this hypothesis.

Evaluations are conducted on the constructed test set. 
For human evaluations, LIMA outperforms InstructGPT and Alpaca by 17\% and 19\%, respectively. Additionally, LIMA achieves comparable results to BARD\footnote{https://bard.google.com/}, Cladue\footnote{https://www.anthropic.com/index/introducing-claude}, and GPT-4. For automatic evaluation, which is conducted by asking GPT-4 to rate responses and a higher rate score denotes better performance, LIMA outperforms InstructGPT and Alpaca by 20\% and 36\%, respectively, achieving comparable results to BARD, while underperforming Claude and GPT-4. Experimental results verify the proposed superficial alignment hypothesis.

\subsection{Others}
\paragraph{OPT-IML (175B)}~\citep{Iyer2022OPTIMLSL} is a large language model trained by fine-tuning the OPT (175B)~\citep{Zhang2022OPTOP} model on the constructed Instruction Meta-Learning (IML) dataset, which consists of over 1500 NLP tasks from 8 publicly available benchmarks such as PromptSource~\citep{Bach2022PromptSourceAI}, FLAN~\citep{longpre2023flan}, and Super-NaturalInstructions~\citep{wang2022super}. After fine-tuning, OPT-IML outperforms OPT across all benchmarks. 

\paragraph{Dolly 2.0 (12B)}~\citep{conover2023free} is initialized with the pre-trained language model Pythia (12B)~\citep{Biderman2023PythiaAS}, and fine-tuned on the instruction dataset databricks-dolly-15k\footnote{https://huggingface.co/datasets/databricks/databricks-dolly-15k}, which contains 7 categories of NLP tasks such as text classification and information extraction. After fine-tuning, Dolly 2.0 (12B) outperforms Pythia (12B) on the EleutherAI LLM Evaluation Harness benchmark~\citep{gao2021framework} by a large margin, and achieves comparable performances to GPT-NEOX (20B)~\citep{Black2022GPTNeoX20BAO}, which has two times more parameters compared to Dolly 2.0 (12B).

\paragraph{Falcon-Instruct (40B)}~\citep{falcon40b} is a large language model trained by fine-tuning Falcon (40B)~\citep{almazrouei2023falcon} on an English dialogue dataset, which contains 150 million tokens from the Baize dataset~\citep{Xu2023BaizeAO}, with an additional 5\% of the data from the RefinedWeb dataset~\citep{penedo2023refinedweb}. To reduce memory usage, the authors employed flash attention~\citep{dao2022flashattention} and multi-query techniques. For evaluation, Falcon-Instruct (40B) achieved better performance on the Open LLM Leaderboard~\citep{beeching2023open}\footnote{https://huggingface.co/spaces/HuggingFaceH4\\/open\_llm\_leaderboard} compared to the baseline model Falcon (40B), and outperforms the Guanaco (65B), which has more model parameters.

\paragraph{Guanaco (7B)}~\citep{Guanaco} is a
multi-turn dialog 
 language model  trained by fine-tuning LLaMA (7B)~\citep{Touvron2023LLaMAOA}  on the constructed multilingual dialogue dataset. The multilingual dialogue dataset comes from two sources:  Alpaca~\citep{taori2023alpaca}, which contains 52K English instruction data pairs; and a multilingual (e.g., Simplified Chinese, Traditional Chinese, Japanese, German) dialogue data, which contains 534K+ multi-turn conversations. 
After fine-tuning, Guanaco 
is to generate role-specific responses and continuous responses on a given topic in multi-turn conversations.

\paragraph{Minotaur (15B)} is a large language model trained by fine-tuning the Starcoder Plus (15B)~\citep{li2023starcoder} on open-source instruction datasets including WizardLM~\citep{xu2023wizardlm} and GPTeacher-General-Instruct\footnote{https://github.com/teknium1/GPTeacher}. For model inference, Minotaur supports a maximum context length of 18K tokens.

\paragraph{Nous-Herme (13B)} is a large language model trained by fine-tuning LLaMA (13B)~\citep{Touvron2023LLaMAOA} on an instruction dataset, which contains over 300k instructions, sampled from  GPTeacher\footnote{https://github.com/teknium1/GPTeacher}, CodeAlpaca~\citep{chaudhary2023code}, GPT-4-LLM~\citep{peng2023instruction}, Unnatural Instructions~\citep{honovich2022unnatural}, and Biology\/Physics\/Chemistry subsets in the Camel-AI~\citep{li2023camel}. Responses are generated by GPT-4. For evaluations, Nous-Herme (13B) achieves comparable performances to GPT-3.5-turbo on multiple tasks like ARC challenge~\citep{Clark2018ThinkYH} and BoolQ~\citep{Clark2019BoolQET}.

\paragraph{TÜLU (6.7B)}~\citep{Wang2023HowFC} is a large language model trained by fine-tuning OPT (6.7B)~\citep{Zhang2022OPTOP} on a mixed instruction dataset, which contains FLAN V2~\citep{longpre2023flan}, CoT~\citep{Wei2022ChainOT}, Dolly~\citep{conover2023free}, Open Assistant-1\footnote{https://huggingface.co/datasets/OpenAssistant/oasst1}, GPT4-Alpaca\footnote{https://huggingface.co/datasets/vicgalle/alpaca-gpt4}, Code-Alpaca~\citep{chaudhary2023code}, and ShareGPT\footnote{https://sharegpt.com/}. After fine-tuning, TÜLU (6.7B) reaches on average 83\% of ChatGPT's performance and 68\% of GPT-4's performance.

\paragraph{YuLan-Chat (13B)}~\citep{YuLan-Chat} is a language model trained by fine-tuning LLaMA (13B)~\citep{Touvron2023LLaMAOA} on a constructed bilingual dataset, which contains 250,000 Chinese-English instruction pairs. After fine-tuning, YuLan-Chat-13B achieves comparable results to the state-of-the-art open-source model ChatGLM (6B)~\citep{du2022glm}, and outperforms Vicuna (13B)~\citep{chiang2023vicuna} on the English BBH3K (BBH3K is a subset of BBH benchmark~\citep{Srivastava2022BeyondTI}) dataset.

\paragraph{MOSS (16B)}\footnote{https://txsun1997.github.io/blogs/moss.html} is a bilingual dialogue language model, which aims to engage in multi-turn conversations and utilize various plugins, trained by fine-tuning on dialogue instructions. After fine-tuning, MOSS outperforms the backbone model and generates responses that better align with human preferences.

\paragraph{Airoboros (13B)}\footnote{https://github.com/jondurbin/airoboros} is a large language model trained by fine-tuning LLAMA (13B)~\citep{Touvron2023LLaMAOA} on the Self-instruct dataset~\citep{wang2022self}. After fine-tuning, Airoboros significantly outperforms LLAMA (13B)~\citep{Touvron2023LLaMAOA} on all benchmarks and achieves highly comparable results to models fine-tuned specifically for certain benchmarks.

\paragraph{UltraLM (13B)}~\citep{ding2023enhancing} is a large language model trained by  fine-tuning LLAMA (13B)~\citep{Touvron2023LLaMAOA}. For evaluation, UltraLM (13B) outperforms Dolly (12B)~\citep{conover2023free} and achieves the winning rate up to 98\%. Additionally, it surpasses the previous best open-source models (i.e., Vicuna~\citep{chiang2023vicuna} and WizardLM~\citep{xu2023wizardlm}) with winning rates of 9\% and 28\%, respectively.

\section{Multi-modality Instruction Tuning} \label{sec:Multi-modality_Instruction_Fine-tuning}

\subsection{Multi-modality Datasets}
\begin{table*}[t]
\centering
\small
\scalebox{0.85}{
\begin{threeparttable}
\begin{tabular}{llll}
\toprule
\multirow{2}{*}{\bf Multi-modality Instruction Fine-tuning Dataset} &  \multicolumn{2}{c}{\bf Modalities} & \multirow{2}{*}{\bf \# Tasks} \\
& {\bf Modality Pair} & {\bf \# Instance} & \\\midrule
{MUL-TIINSTRUCT}~\citep{Xu2022MultiInstructIM}\tnotex{id:1} & Image-Text & 5k to 5M per task & 62 \\
{PMC-VQA}~\citep{Zhang2023PMCVQAVI}\tnotex{id:2} & Image-Text & 227k & 2 \\
\multirow{2}{*}{{LAMM}~\citep{Yin2023LAMMLM}\tnotex{id:3}} & Image-Text & 186k & 9 \\
& Point Cloud-Text & 10k & 3 \\
{Vision-Flan}~\citep{xu2024vision}\tnotex{id:4} & Multi-Pairs & Over 1M & 200+ \\
{ALLAVA}~\citep{chen2024allava}\tnotex{id:5} & Image-Text & 1.4M & 2 \\
{ShareGPT4V}~\citep{chen2023sharegpt4v}\tnotex{id:6} & Image-Text & 1.2M & 2 \\
\bottomrule
\end{tabular}
\end{threeparttable}
}
\begin{multicols}{2}
\begin{tablenotes}
\item[1] \label{id:1} {$^1$ https://github.com/VT-NLP/MultiInstruct} 
\item[2] \label{id:2} {$^2$ https://github.com/xiaoman-zhang/PMC-VQA}
\item[3] \label{id:3} {$^3$ https://github.com/OpenLAMM/LAMM}
\item[4] \label{id:4} {$^4$ https://vision-flan.github.io/}
\item[5] \label{id:5} {$^5$ https://github.com/FreedomIntelligence/ALLaVA}
\item[6] \label{id:6} {$^6$ https://sharegpt4v.github.io/}
\end{tablenotes}
\end{multicols}
\caption{An overview of multi-modality instruction fine-tuning datasets.}
\label{tab:mmllms_dataset_table}
\end{table*}
\paragraph{MUL-TIINSTRUCT}~\citep{Xu2022MultiInstructIM} is a multimodal instruction tuning  dataset consisting of 62 diverse multimodal tasks in a unified seq-to-seq format. This dataset covers 10 broad categories and its tasks are derived from 21 existing open-sourced datasets. 
Each task is equipped with 5 expert-written instructions. For the existing tasks, the authors use the input/output pairs from their available open-source datasets to create instances. While for each new task, the authors create 5k to 5M instances by extracting the necessary information from instances of existing tasks or reformulating them. 
The MUL-TIINSTRUCT dataset has demonstrated its efficiency in enhancing various transfer learning technique. 
For example, fine-tuning the OFA model (930M)~\citep{DBLP:conf/icml/WangYMLBLMZZY22} 
using various transfer learning strategies such as Mixed Instruction Tuning and Sequential Instruction Tuning 
on  MUL-TIINSTRUCT improve the zero-shot performance  across all unseen tasks. On commonsense VQA task, OFA fine-tuned on MUL-TIINSTRUCT achieves 50.60 on RougeL and 31.17 on accuracy, while original OFA achieves 14.97 on RougeL and 0.40 on accuracy.

\paragraph{PMC-VQA}~\citep{Zhang2023PMCVQAVI} is a large-scale medical visual question-answering dataset that comprises 227k image-question pairs of 149k images, covering various modalities or diseases. The dataset can be used for both open-ended and multiple-choice tasks. The pipeline for generating the PMC-VQA dataset involves collecting image-caption pairs from the PMC-OA~\citep{lin2023pmc} dataset, using ChatGPT to generate question-answer pairs, and manually verifying a subset of the dataset for quality. The authors propose a generative-based model MedVInT for medical visual understanding by aligning visual information with a large language model. MedVInT pretrained on PMC-VQA achieves state-of-the-art performance and outperforms existing models on VQA-RAD~\citep{lau2018dataset} and SLAKE~\citep{liu2021slake} benchmarks, with 81.6\% accuracy on VQA-RAD and 88.0\% accuracy on SLAKE.

\paragraph{LAMM}~\citep{Yin2023LAMMLM} is a comprehensive multi-modal instruction tuning dataset for 2D image and 3D point cloud understanding. LAMM contains 186K language-image instruction-response pairs, and 10K language-point cloud instruction-response pairs. The authors collect images and point clouds from publicly available datasets and use the GPT-API and self-instruction methods to generate instructions and responses based on the original labels from these datasets. LAMM-Dataset includes data pairs for commonsense knowledge question answering by incorporating a hierarchical knowledge graph label system from the Bamboo~\citep{zhang2022bamboo} dataset and the corresponding Wikipedia description. The authors also propose the LAMM-Benchmark, which evaluates existing multi-modal language models~(MLLM) on various computer vision tasks. It includes 9 common image tasks and 3 common point cloud tasks, and LAMM-Framework, a primary MLLM training framework that differentiates the encoder, projector, and LLM finetuning blocks for different modalities to avoid modality conflicts. 

\paragraph{Vision-Flan}~\citep{xu2024vision} is the largest public-available human-annotated visual instruction tuning dataset that consists of 1,664,261 instances and 200+ diverse vision-language tasks derived from 101 open-source computer vision datasets. Each task is accompanied by expertly written instructions and meticulously crafted templates for inputs and outputs. The dataset covers a broad spectrum of tasks, including image captioning, visual question-answering, and visual comprehension. Designed to enhance research and application in vision-language model domains, Vision-Flan aims to expand the horizons of interaction and comprehension between visual and linguistic modalities. It provides researchers and developers with a valuable resource to push the envelope of vision-language models and to innovate algorithms across a diverse array of fields.

\paragraph{ALLaVA}\cite{chen2024allava} represents an open-source, extensive dataset tailored for fine-tuning visual question-answering models, featuring 1.4M entries that include detailed captions, intricate instructions, and comprehensive answers produced by GPT-4V \cite{yang2023dawn}. To craft high-quality captions and visual question-answers, \citet{chen2024allava} introduced a method to distill both a caption and a QA pair for an image in a single interaction. This process involves initially presenting GPT-4V \cite{yang2023dawn} with an image, followed by prompting it to generate both a detailed caption and a visual question-answer pair. This approach of incorporating additional visual data enables the model to develop a more nuanced understanding of both the visual and textual elements, enhancing its capacity to deliver precise and contextually appropriate answers. Furthermore, this method has the potential to reduce the occurrence of hallucinations by providing the model with more contextual information (visual data).

\paragraph{ShareGPT4V}\cite{chen2023sharegpt4v} is a collection of highly descriptive image-text pairs, consisting of two components: 100K captions generated by GPT4-Vision \cite{yang2023dawn} from a variety of images, and 1.2M captions developed using their pre-trained model, which was trained on the initial set of 100K high-quality captions. These captions comprehensively cover aspects such as global knowledge, object attributes, spatial relationships, and aesthetic evaluations. Utilizing this dataset, the ShareGPT4V-7B model, once fine-tuned, surpasses competing 7B-scale LMMs across all 11 benchmark tests. Notably, it secures a remarkable cumulative score of 1943.8 on the MME benchmark, outperforming the second-place Qwen-VL-Chat-7B \cite{bai2023qwen} model, which was trained with 1.4 billion samples, by 95.6 points.

\subsection{Multi-modality Instruction Tuning Models}
\begin{figure}[t]
  \centering
  \begin{minipage}[t]{0.5\textwidth}
    \centering
    \includegraphics[width=1\textwidth]{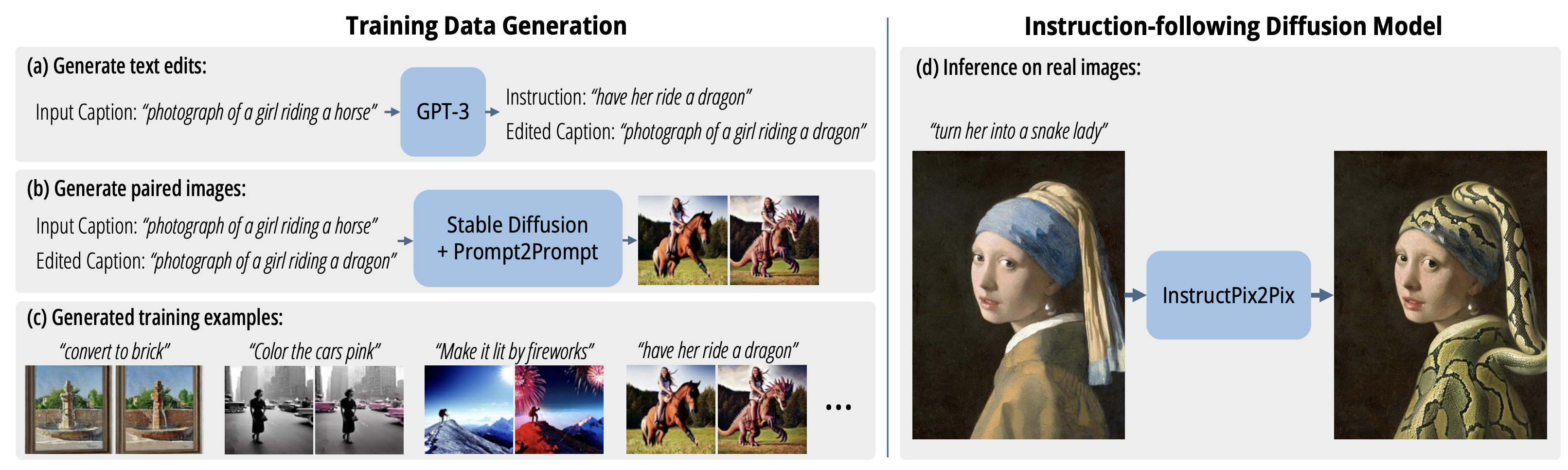}
  \end{minipage}%
  \caption{Image editing dataset generation and diffusion model training. The figure is copied from \citet{Brooks2022InstructPix2PixLT}.}
  \label{fig:InstructPix2Pix}
\end{figure}

\begin{table*}[t]
\centering
\small
\scalebox{0.85}{
\begin{threeparttable}
\begin{tabular}{lllllll}
\toprule
{\bf Multi-modality Instruction} & \multirow{2}{*}{\bf \# Params} & \multirow{2}{*}{\bf Modality} & \multicolumn{2}{c}{\bf Base Model} & \multicolumn{2}{c}{\bf Fine-tuning Trainset}\\
{\bf Fine-tuned LLMs} & & & {\bf Model Name} & {\bf \# Params} & {\bf Self-build} & {\bf Size} \\\midrule
{{InstructPix2Pix}~\citep{Brooks2022InstructPix2PixLT}\tnotex{id:1}} & 983M & I/T & Stable Diffusion & 983M & Yes & 450K \\
\midrule
\multirow{2}{*}{{LLaVA}~\citep{Liu2023VisualIT}\tnotex{id:2}} & \multirow{2}{*}{13B} & \multirow{2}{*}{I/T} & CLIP~\citep{Radford2021LearningTV} & 400M & \multirow{2}{*}{Yes} & 158K \\
& & & LLaMA~\citep{Touvron2023LLaMAOA} & 7B & & \\
& & & LLaMA~\citep{Touvron2023LLaMAOA} & 7B & & \\
\midrule
\multirow{3}{*}{Video-LLaMA~\citep{damonlpsg2023videollama}\tnotex{id:3}} & \multirow{3}{*}{-} & \multirow{3}{*}{I/T/V/A} & BLIP-2~\citep{li2023blip2} & - & \multirow{3}{*}{No} & \multirow{3}{*}{-} \\
& & & ImageBind~\citep{girdhar2023imagebind} & - & & \\
& & & Vicuna~\citep{chiang2023vicuna} & 7B/13B & & \\
\midrule
{InstructBLIP (1.2B)}~\citep{Dai2023InstructBLIPTG}\tnotex{id:4} & - & I/T/V & BLIP-2~\citep{li2023blip2} & - & No & - \\
\midrule
{Otter}~\citep{Li2023OtterAM}\tnotex{id:5} & - & I/T/V & OpenFlamingo~\citep{anas_awadalla_2023_7733589} & 9B & Yes & 2.8M \\
\midrule
{MultiModal-GPT}~\citep{Gong2023MultiModalGPTAV}\tnotex{id:6} & - & I/T/V & OpenFlamingo~\citep{anas_awadalla_2023_7733589} & 9B & No & - \\
\bottomrule
\end{tabular}
\end{threeparttable}
}
\begin{multicols}{2}
\begin{tablenotes}
\item[1] \label{id:1} {$^1$ https://github.com/timothybrooks/instruct-pix2pix} 
\item[2] \label{id:2} {$^2$ https://github.com/haotian-liu/LLaVA}
\item[3] \label{id:3} {$^3$ https://github.com/DAMO-NLP-SG/Video-LLaMA}
\item[4] \label{id:4} {$^4$ https://github.com/salesforce/LAVIS/tree/main/projects/instructblip}
\item[5] \label{id:5} {$^5$ https://github.com/Luodian/Otter}
\item[6] \label{id:6} {$^6$ https://github.com/open-mmlab/Multimodal-GPT}
\end{tablenotes}
\end{multicols}
\caption{An overview of multi-modality instruction fine-tuned LLMs. I/T/V/A stand for Image/Text/Video/Audio}
\label{tab:mmllms_model_table}
\end{table*}
\paragraph{InstructPix2Pix (983M)}~\citep{Brooks2022InstructPix2PixLT} is a conditional diffusion model trained by fine-tuning Stable Diffusion (983M)~\citep{Rombach_2022_CVPR} on a constructed multi-modal dataset that contains more than 450K text editing instructions and corresponding images before and after the edit. The authors combine the abilities of two large-scale pre-trained models, a language model GPT-3~\citep{Brown2020LanguageMA} and a text-to-image model Stable Diffusion~\citep{Rombach_2022_CVPR}, to generate the the training dataset. GPT-3 is fine-tuned to generate text edits based on image prompts, while Stable Diffusion is used to convert the generated text edits into actual image edits. InstructPix2Pix is then trained on this generated dataset using a latent diffusion objective. Figure~\ref{fig:InstructPix2Pix} shows the process of generating image editing dataset and training the diffusion model on that dataset. The authors compares the proposed method qualitatively with previous works such as SDEdit~\citep{meng2022sdedit} and Text2Live~\citep{bar2022text2live}, highlighting the ability of the model to follow image editing instructions instead of descriptions of the image or edit layer. The authors also presents quantitative comparisons with SDEdit~\citep{meng2022sdedit} using metrics measuring image consistency and edit quality. 

\paragraph{LLaVA (13B)}~\citep{Liu2023VisualIT} is a large multimodal model developed by connecting the visual encoder of CLIP (400M)~\citep{Radford2021LearningTV} with the language decoder LLaMA (7B)~\citep{Touvron2023LLaMAOA}. LLaVA is fine-tuned using the generated instructional vision-language dataset consisted of 158K unique language-image instruction-following samples. The data collection process involved creating conversation, detailed description, and complex reasoning prompts. GPT-4 is used to convert image-text pairs into appropriate instruction-following format for this dataset. Visual features such as captions and bounding boxes were used to encode images. LLaVA yields a 85.1\% relative score compared with GPT-4 on a synthetic multimodal instruction following dataset. When fine-tuned on Science QA, the synergy of LLaVA and GPT-4 achieves a new state-of-the-art accuracy of 92.53\%.

\begin{figure}[t]
  \centering
  \begin{minipage}[t]{0.5\textwidth}
    \centering
    \includegraphics[width=1\textwidth]{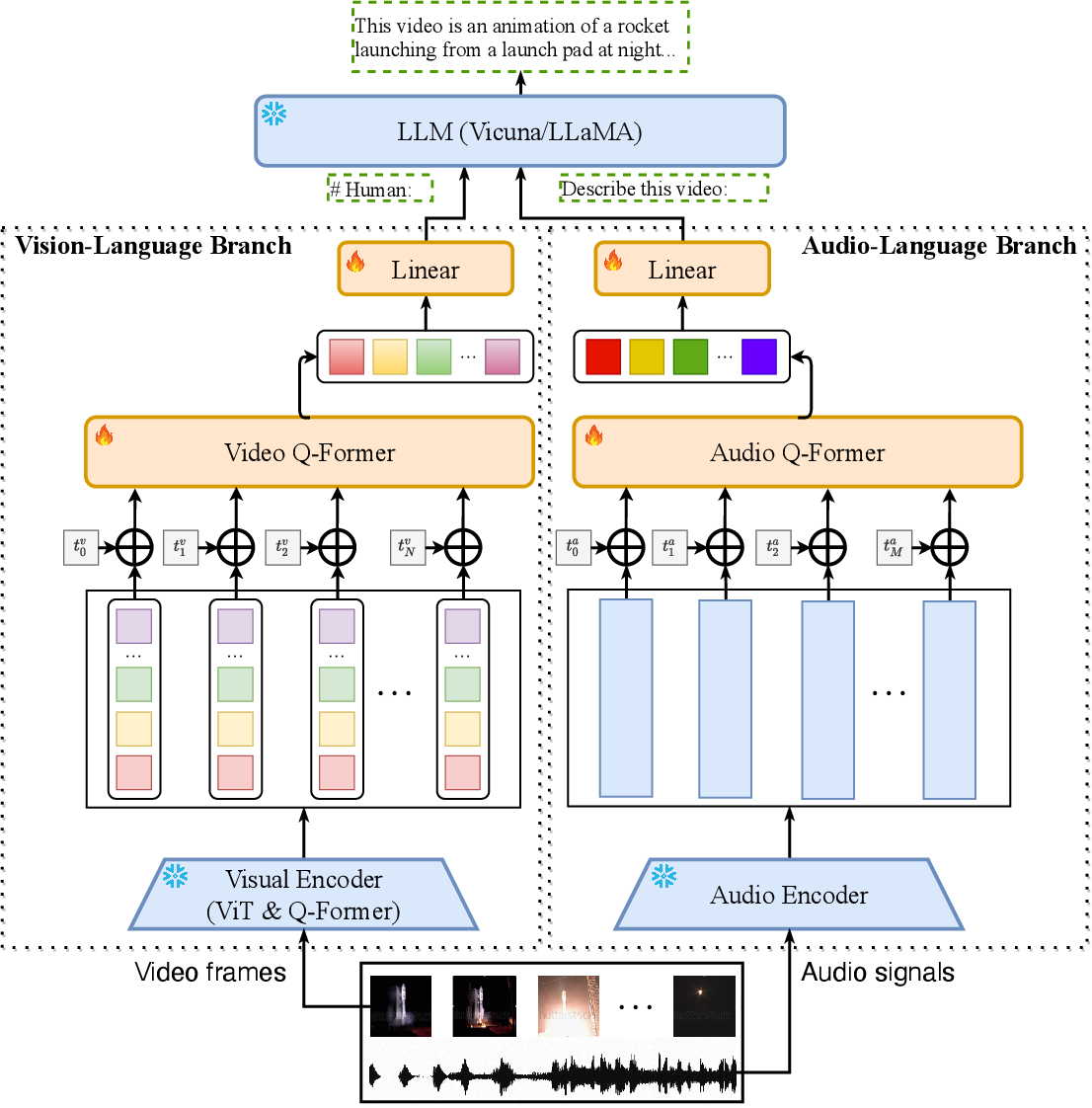}
  \end{minipage}%
  \caption{Overall architecture of Video-LLaMA. The figure is copied from \citet{damonlpsg2023videollama}.}
  \label{fig:Video-LLaMA}
\end{figure}

\paragraph{Video-LLaMA}~\citep{damonlpsg2023videollama} is a multimodal framework that enhances large language models with the ability to understand both visual and auditory content in videos. The architecture of Video-LLaMA consists of two branche encoders: the Vision-Language~(VL) Branch and the Audio-Language~(AL) Branch, and a language decoder (Vicuna (7B/13B)~\citep{chiang2023vicuna}, LLaMA (7B)~\citep{Touvron2023LLaMAOA}, etc.). 
The VL Branch includes a frozen pre-trained image encoder (pre-trained vision component of BLIP-2~\citep{li2023blip2}, which includes a ViT-G/14 and a pre-trained Q-former), a position embedding layer, a video Q-former and a linear layer. 
The AL Branch includes a pre-trained audio encoder (ImageBind~\citep{girdhar2023imagebind}) and an Audio Q-former. Figure~\ref{fig:Video-LLaMA} shows the overall architecture of Video-LLaMA with Vision-Language Branch and Audio-Language Branch. 
The VL Branch is trained on the Webvid-2M~\citep{bain2021frozen} video caption dataset with a video-to-text generation task, and fine-tuned on the instruction tuning data from MiniGPT-4~\citep{zhu2023minigpt}, LLaVA~\citep{Liu2023VisualIT} and VideoChat~\citep{2023videochat}. The AL Branch is trained on video/image instru-caption data to connect the output of ImageBind to language decoder. 
After finetuning, Video-LLaMA can perceive and comprehend video content, demonstrating its ability to integrate auditory and visual information, understand static images, recognize common-knowledge concepts, and capture temporal dynamics in videos. 

\begin{figure}[t]
  \centering
  \begin{minipage}[t]{0.5\textwidth}
    \centering
    \includegraphics[width=1\textwidth]{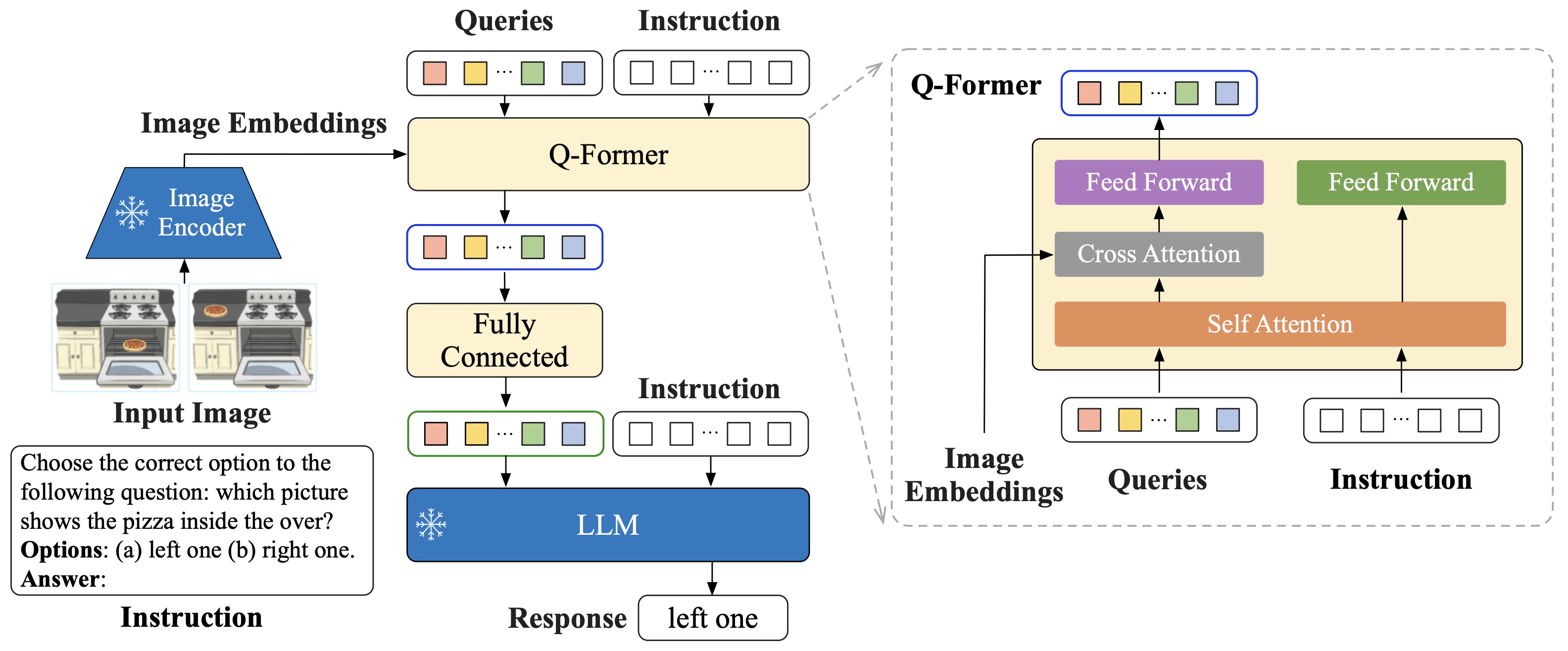}
  \end{minipage}%
  \caption{Overall architecture of InstructBLIP. The figure is copied from \citet{Dai2023InstructBLIPTG}.}
  \label{fig:InstructBLIP}
\end{figure}

\paragraph{InstructBLIP (1.2B)}~\citep{Dai2023InstructBLIPTG} is a vision-language instruction tuning framework initialized with a pre-trained BLIP-2~\citep{li2023blip2}) model consisting of an image encoder, an LLM (FlanT5 (3B/11B)~\citep{Chung2022ScalingIL} or Vicuna (7B/13B)~\citep{chiang2023vicuna}), and a Query Transformer (Q-Former) to bridge the two. As shown in Figure \ref{fig:InstructBLIP}, the Q-Former extracts instruction-aware visual features from the output embeddings of the frozen image encoder, and feeds the visual features as soft prompt input to the frozen LLM. The authors evaluate the proposed InstructBLIP model on a variety of vision-language tasks, including image classification, image captioning, image question answering, and visual reasoning. They use 26 publicly available datasets, dividing them into 13 held-in and 13 held-out datasets for training and evaluation. The authors demonstrate that InstructBLIP achieves state-of-the-art zero-shot performance on a wide range of vision-language tasks. InstructBLIP yields an average relative improvement of 15.0\% when compared to BLIP-2, smallest InstructBLIP (4B) outperforms Flamingo (80B)~\citep{DBLP:conf/nips/AlayracDLMBHLMM22} on all six shared evaluation datasets with an average relative improvement of 24.8\%.

\paragraph{Otter}~\citep{Li2023OtterAM} is a multi-modal model trained by fine-tuning OpenFlamingo (9B)~\citep{anas_awadalla_2023_7733589}, with the language and vision encoders frozen and only fine-tuning the Perceiver resampler module, cross-attention layers, and input/output embeddings. The authors organize diverse multi-modal tasks covering 11 categories and build multi-modal in-context instruction tuning datasets MIMIC-IT of 2.8M multimodal instruction-response pairs, which consists of image-instruction-answer triplets, where the instruction-answer is tailored to the image. Each data sample also includes context, which contains a series of image-instruction-answer triplets that contextually correlate with the queried triplet. Otter demonstrates the ability to follow user instructions more accurately and provide more detailed descriptions of images compared to OpenFlamingo~\citep{anas_awadalla_2023_7733589}. 

\paragraph{MultiModal-GPT}~\citep{Gong2023MultiModalGPTAV} is a multi-modal instruction tuning model that is capable of following diverse instructions, generating detailed captions, counting specific objects, and addressing general inquiries. MultiModal-GPT is trained by fine-tuning OpenFlamingo (9B)~\citep{anas_awadalla_2023_7733589} on various created visual instruction data with open datasets, including VQA, Image Captioning, Visual Reasoning, Text OCR, and Visual Dialogue. The experiments demonstrate the proficiency of MultiModal-GPT in maintaining continuous dialogues with humans.

\section{Domain-specific Instruction Tuning}
In this section, we describe instruction tuning in different domains and applications.
\begin{table*}[t]
\centering
\small
\scalebox{0.85}{
\begin{threeparttable}
\begin{tabular}{lllll}
\toprule
 \multirow{1}{*}{\bf Domain Type} &  {\bf Domain-specific Instruction} & \multicolumn{2}{c}{\bf Base Model} & \multirow{2}{*}{\bf Trainset Size}\\
& {\bf Fine-tuned LLMs} & {\bf Model Name} & {\bf \# Params} \\\midrule
 \multirow{1}{*}{Dialogue} & \multirow{1}{*}{{InstructDial}~\citep{Gupta2022InstructDialIZ}\tnotex{id:1}}  & T0~\citep{sanh2021multitask} & 3B & \multirow{2}{*}{-} \\
Classification & {{LINGUIST}~\citep{Rosenbaum2022LINGUISTLM}}  & AlexaTM~\citep{soltan2022alexatm} & 5B &  13K \\
Information extraction & {InstructUIE}~\citep{Wang2023InstructUIEMI}\tnotex{id:2} & FlanT5~\citep{Chung2022ScalingIL} & 11B & 1.0M \\

Sentiment analysis & IT-MTL~\citep{Varia2022InstructionTF}\tnotex{id:3} &  T5~\citep{Raffel2019ExploringTL} & 220M & - \\
\midrule
\multirow{3}{*}{\shortstack{ Writing}}   & {Writing-Alpaca-7B}~\citep{Zhang2023MultiTaskIT}\tnotex{id:4} & LLaMA~\citep{Touvron2023LLaMAOA} & 7B & - \\

& \multirow{1}{*}
{{CoEdIT}~\citep{Raheja2023CoEdITTE}\tnotex{id:5}}  & FlanT5~\citep{Chung2022ScalingIL} & 11B & \\

& \multirow{1}{*}
{{CoPoet}~\citep{Chakrabarty2022HelpMW}\tnotex{id:6}}  & T5~\citep{Raffel2019ExploringTL} & 11B & \\
\midrule 
\multirow{3}{*}{\shortstack{ Medical}}   & {Radiology-GPT}~\citep{Liu2023RadiologyGPTAL}\tnotex{id:7} & Alpaca~\citep{taori2023alpaca} & 7B &  122K \\

 & {ChatDoctor}~\citep{Li2023ChatDoctorAM}\tnotex{id:8} & LLaMA~\citep{Touvron2023LLaMAOA} & 7B &  100K \\
 & {ChatGLM-Med}~\citep{ChatGLM-Med}\tnotex{id:9} & ChatGLM~\citep{du2022glm} & 6B & - \\
\midrule
Arithmetic& {Goat}~\citep{liu2023goat}\tnotex{id:10}  & LLaMA~\citep{Touvron2023LLaMAOA} & 7B & 1.0M \\

Code & {WizardCoder}~\citep{luo2023wizardcoder}\tnotex{id:11}  & StarCoder~\citep{li2023starcoder} & 15B &  78K \\
\bottomrule
\end{tabular}
\end{threeparttable}
}
\begin{multicols}{2}
\begin{tablenotes}
\item[1] \label{id:1} {$^1$ https://github.com/prakharguptaz/Instructdial} 
\item[2] \label{id:2} {$^2$ https://github.com/BeyonderXX/InstructUIE}
\item[3] \label{id:3} {$^3$ https://github.com/amazon-science/instruction-tuning-for-absa}
\item[4] \label{id:4} {$^4$ https://github.com/facebookresearch/EditEval}
\item[5] \label{id:5} {$^5$ https://github.com/vipulraheja/coedit}
\item[6] \label{id:6} {$^6$ https://github.com/vishakhpk/creative-instructions}
\item[7] \label{id:7} {$^7$ https://huggingface.co/spaces/allen-eric/radiology-gpt}
\item[8] \label{id:8} {$^8$ https://github.com/Kent0n-Li/ChatDoctor}
\item[9] \label{id:9} {$^9$ https://github.com/SCIR-HI/Med-ChatGLM}
\item[10] \label{id:10} {$^{10}$ https://github.com/liutiedong/goat}
\item[11] \label{id:11} {$^{11}$ https://github.com/nlpxucan/WizardLM}
\end{tablenotes}
\end{multicols}
\caption{An overview of domain-specific instruction fine-tuned LLMs. }
\label{tab:mmllms_model_table}
\end{table*}
\label{application}
\subsection{Dialogue}
\paragraph{InstructDial}~\citep{Gupta2022InstructDialIZ} is an instruction tuning framework designed for dialogue. It contains a collection of 48 dialogue tasks in a consistent text-to-text format created from 59 dialogue datasets. Each task instance includes a task description, instance inputs, constraints, instructions, and output. To ensure adherence to instructions, the framework introduces two  meta-tasks: (1) an instruction selection task, where the model selects the instruction corresponding to a given input-output pair; 
and (2) an instruction binary task, where the model predicts "yes" or "no" if an instruction leads to a given output from an input. Two base models T0-3B~\citep{sanh2021multitask} (3B parameters version of T5~\citep{lester-etal-2021-power}) and BART0~\citep{lin2022unsupervised} (406M parameters based on Bart-large~\citep{lewis-etal-2020-bart}) are fine-tuned on the tasks from InstructDial. InstructDial achieves impressive results on unseen dialogue datasets and tasks, including dialogue evaluation and intent detection. Moreover, it delivers even better results when applied to a few-shot setting.

\subsection{Intent Classification and Slot Tagging}
\paragraph{LINGUIST}~\citep{Rosenbaum2022LINGUISTLM} finetunes AlexaTM 5B~\citep{soltan2022alexatm}, a 5-billion-parameter multilingual model, on the instruction dataset for intent classification and slot tagging tasks. Each instruction consists of five blocks: (i) the language of the generated output, (ii) intention, (iii) slot types and values to include in the output (e.g., the number 3 in [3, snow] corresponds the slot type, and snow is the value used for that slot), (iv) a mapping from slot type labels to numbers, and (v) up to 10 examples to instruct the format of the outputs.  LINGUIST  shows significant improvements over state-of-the-art approaches in a 10-shot novel intent setting using the SNIPS dataset~\citep{DBLP:journals/corr/abs-1805-10190}. In the zero-shot cross-lingual setting of the mATIS++ dataset~\citep{xu2020endtoend}, LINGUIST surpasses a strong baseline of Machine Translation with Slot Alignment across 6 languages while maintaining intent classification performance.

\begin{figure}[t]
  \centering
  \begin{minipage}[t]{0.5\textwidth}
    \centering
    \includegraphics[width=1\textwidth]{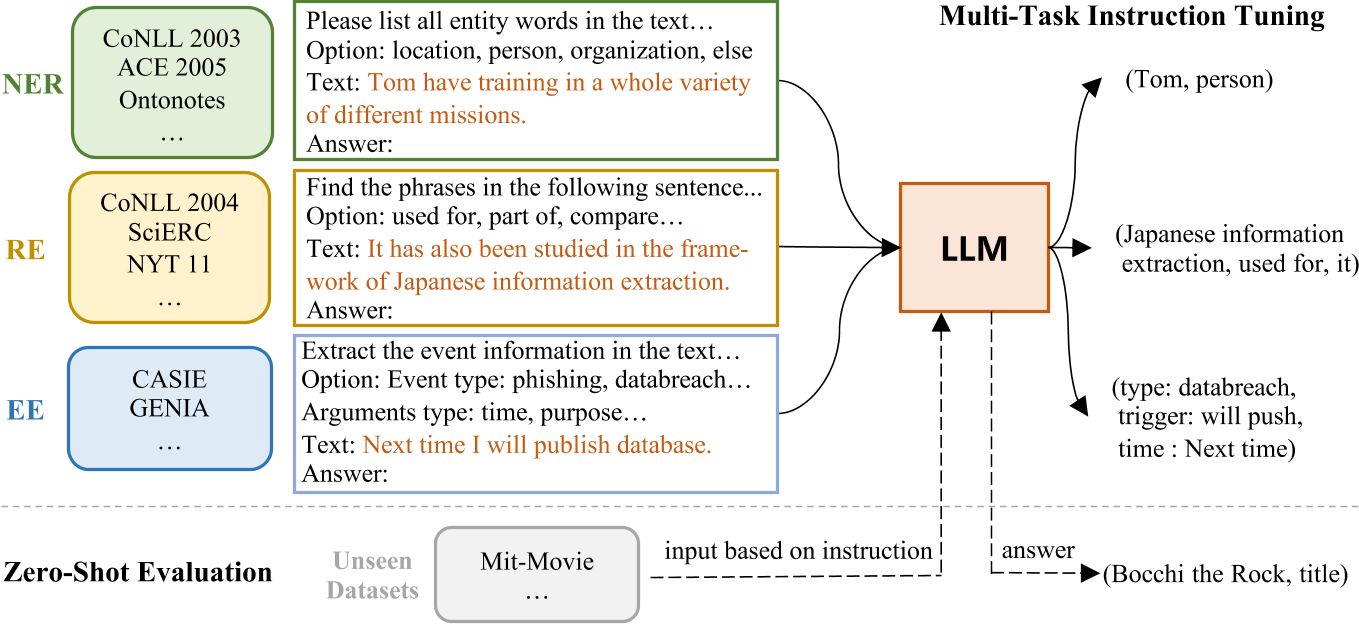}
  \end{minipage}%
  \caption{The overview framework of InstructUIE. The figure is copied from \citet{Wang2023InstructUIEMI}.}
  \label{fig:InstructUIE}
\end{figure}

\subsection{Information Extraction}
\paragraph{InstructUIE}~\citep{Wang2023InstructUIEMI} is a unified information extraction (IE) framework based on instruction tuning, which transforms IE tasks to the seq2seq format and solves them by fine-tuning 11B FlanT5~\citep{Chung2022ScalingIL} on the constructed SFT dataset. Figure~\ref{fig:InstructUIE} shows the overall architecture of InstructUIE. It introduces IE INSTRUCTIONS, a benchmark of 32 diverse information extraction datasets in a unified text-to-text format with expert-written instructions. Each task instance is delineated by four properties: task instruction, options, text, and output. Task instruction contains information such as the type of information to be extracted, the output structure format, and additional constraints or rules that need to be adhered to during the extraction process. Options refer to the output label constraints of a task.
Text refers to the input sentence. Output is the sentence obtained by  converting the original tags of the sample (e.g. "entity tag: entity span" for NER). In the supervised setting, InstructUIE performs comparably to BERT~\citep{devlin-etal-2019-bert} and outperforms the state-of-the-art and GPT3.5~\citep{brown2020language} in zero-shot settings.

\subsection{Aspect-based Sentiment Analysis}
\paragraph{\citet{Varia2022InstructionTF}} propose a unified instruction tuning framework for solving Aspect-based Sentiment Analysis (ABSA) task based on a fine-tuned T5 (220M)~\citep{Raffel2019ExploringTL} model. The framework addresses multiple factorized sub-tasks that involve the four elements of ABSA, namely Aspect Term, Aspect Category, Opinion Term, and Sentiment. It treats these sub-tasks as a combination of five Question Answering (QA) tasks by transforming each sentence in the corpus using instruction templates provided for each task. For instance, one of the instruction templates used is "What are the aspect terms in the text: \$TEXT?". The framework showcases substantial improvement (8.29 F1 on average) over the state-of-the-art in few-shot learning scenarios and remains comparable in full fine-tuning scenarios.

\subsection{Writing}
\paragraph{\citet{Zhang2023MultiTaskIT}} propose Writing-Alpaca-7B that fine-tunes LLaMa-7B~\citep{peng2023instruction} on the writing  instruction dataset to provide writing assistance. The proposed instruction dataset is an extension of the EDITEVAL~\citep{dwivediyu2022editeval} benchmark based on instructional data, with the Updating task removed and a task for grammaticality introduced. The instruction scheme strictly follows the one in the Stanford Alpaca project~\citep{taori2023alpaca}, comprising a universal preface, an instruction field to guide task completion, an input field that provides the text to be edited, and a response field that requires models to fill out. The Writing-Alpaca-7B improves upon LLaMa’s performance on all writing tasks and outperforms other larger off-the-shelf LLMs.

\begin{figure}[t]
  \centering
  \begin{minipage}[t]{0.5\textwidth}
    \centering
    \includegraphics[width=1\textwidth]{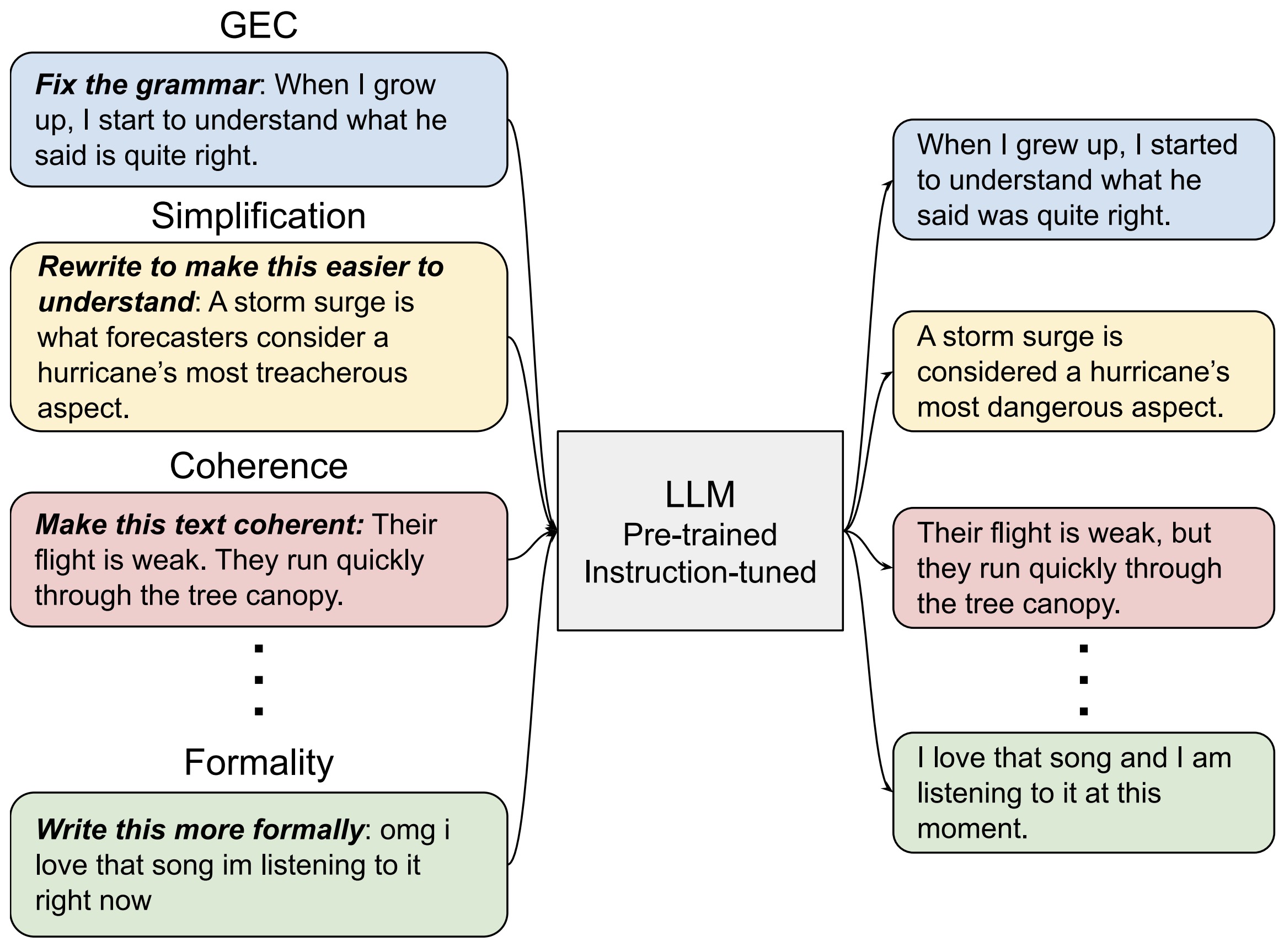}
  \end{minipage}%
  \caption{The overview framework of COEDIT. The figure is copied from \citet{Raheja2023CoEdITTE}.}
  \label{fig:COEDIT}
\end{figure}

\paragraph{CoEdIT}~\citep{Raheja2023CoEdITTE} finetunes FLANT5~\citep{Chung2022ScalingIL} (770M parameters, 3B parameters, and 11B parameters) on the instruction dataset for text editing to provide writing assistance. The instruction dataset comprises approximately 82K <instruction: source, target> pairs. As shown in Figure~\ref{fig:COEDIT}, the model takes instructions from the user specifying the characteristics of the desired text, such as "Make the sentence simpler", and outputs the edited text. CoEdIT achieves state-of-the-art performance on several text editing tasks, including grammatical error correction, text simplification, iterative text editing, and three stylistic editing tasks: formality style transfer, neutralization, and paraphrasing. Furthermore, it can generalize well to new, adjacent tasks not seen during fine-tuning.

\paragraph{CoPoet}~\citep{Chakrabarty2022HelpMW} is a collaborative poetry writing tool that utilizes a large language model (e.g. T5-3B, T5-11B and T0-3B models) trained on a diverse collection of instructions for poetry writing. Each sample in the instruction dataset includes an <instruction, poem\_line> pair. There are three major types of instructions: Continuation, Lexical Constraints, and Rhetorical Techniques. The CoPoet is guided by user instructions that specify desired attributes of the poetry, such as writing a sentence about "love" or ending a sentence with "fly." Not only is the system competitive with publicly available LLMs trained on instructions, such as InstructGPT~\citep{ouyang2022training}, but it is also capable of satisfying unseen compositional instructions.

\subsection{Medical}
\paragraph{Radiology-GPT}~\citep{Liu2023RadiologyGPTAL} is a fine-tuned Alpaca-7B~\citep{taori2023alpaca} model for radiology, which utilizes an instruction tuning approach on an extensive dataset of radiology domain knowledge. Radiology reports usually include two corresponding sections: "Findings" and "Impression". The "Findings" section contains detailed observations from the radiology images, while the "Impression" section summarizes the interpretations drawn from those observations. Radiology-GPT provides a brief instruction to the "Findings" text: "Derive the impression from findings in the radiology report". The "Impression" text from the same report serves as the target output. In comparison to general language models such as StableLM~\citep{stabilityStabilityLaunches}, Dolly~\citep{conover2023free}, and LLaMA~\citep{Touvron2023LLaMAOA}, Radiology-GPT demonstrates significant adaptability in radiological diagnosis, research, and communication.

\paragraph{ChatDoctor}~\citep{Li2023ChatDoctorAM} is based on the fine-tuned LLaMa-7B~\citep{Touvron2023LLaMAOA} model, utilizing the alpaca instruction dataset~\citep{taori2023alpaca} and the HealthCareMagic100k patient-doctor dialogue dataset. And prompt templates are designed for retrieving external knowledge databases, such as the Disease Database and Wikipedia retrieval, during doctor-patient conversations to obtain more accurate outputs from the model. The ChatDoctor significantly improves the model's ability to comprehend patient needs and provide informed advice. By equipping the model with self-directed information retrieval from reliable online and offline sources, the accuracy of its responses is substantially improved.

\paragraph{ChatGLM-Med}~\citep{ChatGLM-Med} is fine-tuned on the Chinese medical instruction dataset based on the ChatGLM-6B~\citep{du2022glm} model. The instruction dataset comprises medically relevant question and answer pairs, created using the GPT 3.5 API and the Medical Knowledge Graph. This model improves the question-answering performance of ChatGLM~\citep{du2022glm} in the medical field.

\subsection{Arithmetic}
\paragraph{Goat}~\citep{liu2023goat} is a fine-tuned LLaMA-7B~\citep{Touvron2023LLaMAOA} model based on instructions, which aims to solve arithmetic problems. It expresses arithmetic problems in the form of natural language question answering, such as "What is 8914/64?", by generating hundreds of instruction templates using ChatGPT~\citep{chatgpt}. The model applies various techniques to enhance its adaptability to diverse question formats, such as randomly removing spaces between numbers and symbols in the arithmetic expression and replacing "*" with "x" or "times". The Goat model achieves state-of-the-art performance on the BIG-bench~\citep{Srivastava2022BeyondTI} arithmetic subtask. In particular, zero-shot Goat-7B matches or exceeds the accuracy achieved by the few-shot PaLM-540B~\citep{Chowdhery2022PaLMSL}.

\subsection{Code}
\paragraph{WizardCoder}~\citep{luo2023wizardcoder} utilizes StarCoder 15B~\citep{li2023starcoder} as the foundation with complex instruction tuning, by adapting the Evol-Instruct method~\citep{xu2023wizardlm} to the domain of code. The training dataset is produced through iterative application of the Evol-Instruct technique on the Code Alpaca dataset~\citep{alpaca}, which includes the following attributes for each sample: instruction, input, and expected output. For instance, when the instruction is "Amend the following SQL query to select distinct elements", the input is the SQL query, and the expected output is the generated answer. The WizardCoder outperforms all other open-source Code LLMs and even surpasses the largest closed LLMs, Anthropic's Claude and Google's Bard, on HumanEval and HumanEval+.

\section{Efficient Tuning Techniques} \label{Efficient_fine-tuning_techniques}

Efficient fine-tuning techniques aim at adapting LLMs to downstream tasks by optimizing a small fraction of parameters in multiple ways, \textit{i.e.}, addition-based, specification-based, and reparameterization-based.
Addition-based methods introduce extra trainable parameters or modules not present in the original model. Representative methods include adapter tuning~\citep{DBLP:conf/icml/HoulsbyGJMLGAG19} and prompt-based tuning~\citep{DBLP:conf/eacl/SchickS21}. Specification-based methods specify certain inherent model parameters to be tuned while freezing others. For example, BitFit~\citep{DBLP:conf/acl/ZakenGR22} tunes the bias terms of the pre-trained model. Reparameterization methods transform model weights into more parameter-efficient forms for tuning. The key hypothesis is that model adaptation is low-rank, so weights can be reparameterized into low-rank factors or a low-dimensional subspace (\textit{e.g.}, LoRA~\citep{hu2021lora}). Intrinsic prompt tuning finds a low-dimensional subspace shared by tuning prompts across diverse tasks. 
\subsection{LoRA}
 Low-Rank Adaptation (LoRA) \cite{hu2021lora} enables efficient adaptation of LLMs using low-rank updates. LoRA use DeepSpeed~\citep{rasley2020deepspeed} as the training backbone.
     The key insight of LoRA is that the actual change in LLMs' weights required for new task adaptation lies in a low-dimensional subspace. 
    Specifically, for a pretrained weight matrix $W_0$, the authors model the adapted weight matrix as $W_0$ + $\Delta W$, where $\Delta W$ is a low rank update. $\Delta W$ is parameterized as $\Delta W = BA$, where $A$ and $B$ are much smaller trainable matrices. The rank $r$ of $\Delta W$ is chosen to be much smaller than the dimensions of $W_0$. 
    The intuition is that instead of directly training all of $W_0$, the authors train low-dimensional $A$ and $B$, which indirectly trains $W_0$ in a low-rank subspace of directions that matter for the downstream task. This results in far fewer trainable parameters compared to full fine-tuning.
    For GPT-3, LoRA reduces the number of trainable parameters by 10,000x and memory usage by 3x compared to full fine-tuning.
    \subsection{HINT}
HINT \cite{Ivison2022HINTHI} combines the generalization benefits of instruction tuning with efficient on-demand fine-tuning, avoiding repeatedly processing lengthy instructions.
The essence of HINT lies in hypernetworks, which generate parameter-efficient modules for LLMs adaptation based on natural language instructions and few-shot examples. The adopted hypernetwork converts instructions and few-shot examples into a encoded instruction and generates adapter and prefix parameters using a pretrained text encoder and cross-attention based parameter generator. Then, the generated adapters and prefixes are inserted into the backbone model as efficient tuning modules. At inference, the hypernetwork performs inference only once per task to generate adapted modules.
The benefits are that HINT can incorporate long instructions and additional few-shots without increasing compute, unlike regular fine-tuning or input concatenation methods.
    \subsection{Qlora}
QLORA \cite{dettmers2023qlora} includes optimal quantization and memory optimization, aiming at providing efficient and effective LLMs fine-tuning.
QLORA includes 4-bit NormalFloat (NF4) Quantization, which is a quantization scheme optimized for the typical normal distribution of LLM weights. By quantizing based on the quantiles of a normal distribution, NF4 provides better performance than standard 4-bit integer or float quantization. To further reduce memory, the quantization constants are themselves quantized to 8 bits. This second level of quantization saves an additional 0.37 bits per parameter on average. QLORA leverages NVIDIA's unified memory feature to page optimizer states to CPU RAM when GPU memory is exceeded. avoiding out-of-memory during training.
QLORA enables training a 65B parameter LLM on a single 48GB GPU with no degradation compared to full 16-bit finetuning. QLORA works by freezing the 4-bit quantized base LLM, then backpropagating through it into a small set of 16-bit low-rank adapter weights which are learned.
    \subsection{LOMO}
LOw-Memory Optimization (LOMO) \cite{Lv2023FullPF} enables full parameter fine-tuning of LLMs using limited computational resources through a fusion of gradient computation and update. The essence is to fuse gradient computation and parameter update into one step during backpropagation, thereby avoiding storage of full gradient tensors. Firstly, theoretical analysis is provided in LOMO on why SGD can work well for fine-tuning large pre-trained models despite its challenges on smaller models. In addition, LOMO updates each parameter tensor immediately after computing its gradient in backpropagation. Storing the gradient of one parameter at a time reduces gradient memory to $O(1)$. LOMO employs gradient value clipping, separate gradient norm computation pass and dynamic loss scaling to stabilize training. The integration of activation checkpointing and ZeRO optimization methods saves memory.
    
    \subsection{Delta-tuning}
Delta-tuning \cite{Ding2023ParameterefficientFO} provides optimization and optimal control perspectives for theoretical analyzation. Intuitively, delta-tuning performs subspace optimization by restricting tuning to a low-dimensional manifold. The tuned parameters act as optimal controllers guiding model behavior on downstream tasks.

\section{Evaluation, Analysis and Criticism}
\label{analysis}

\subsection{Close-ended Evaluations}
It is widely accepted among researchers that general-purpose models must demonstrate proficiency in certain core tasks before they can effectively generalize to meet diverse real-world needs. Close-ended evaluations help achieve this objective, often involving multiple-choice questions to assess the performance of LLMs.
Below are 6 widely used close-ended evaluations:

\paragraph{(1) MMLU.} 
Massive Multitask Language Understanding (MMLU) \cite{hendrycks2020measuring} consists of 14079 questions covering 57 tasks including elementary mathematics, US history, computer science, law, and more. The wide range of subjects and complex questions make MMLU suitable for testing the model's language comprehension and decision-making capabilities.

\paragraph{(2) MATH and (3) GSM8K.}
MATH \cite{hendrycks2021measuring} and GSM8K \cite{cobbe2021training} are two distinct mathematical datasets utilized for evaluating different aspects of model capabilities. The MATH \cite{hendrycks2021measuring} dataset comprises 12,500 complex competition-level mathematics problems, primarily designed to assess the ability of models to tackle challenging and advanced mathematical questions typically encountered at the college level. Conversely, the GSM8K \cite{cobbe2021training} dataset contains 8,500 high-quality elementary school math problems, aimed at testing the basic mathematical reasoning abilities of models.

\paragraph{(4) BBH.}
BBH, short for BIG-Bench Hard \cite{suzgun2022challenging}, is a subset of the BIG-Bench \cite{srivastava2022beyond} dataset comprising 23 challenging tasks. These tasks were selected because they consistently proved too difficult for current large language models to handle effectively. Requiring complex, multi-step reasoning, the BBH dataset is primarily utilized to assess the general reasoning capabilities of models, testing their ability to navigate and solve intricate problems.

\paragraph{(5) HumanEval (Coding).}
HumanEval \cite{chen2021evaluating} consists of 164 programming problems, including language comprehension, algorithms, and simple mathematics, with some comparable to simple software interview questions. 
The primary purpose of this dataset is to assess the ability of models to generate correct programs based on provided docstrings.

\paragraph{(6) IFEval.}
IFEval \cite{zhou2023instruction} consists of 500 prompts, each containing specific instructions like "write an article with more than 800 words" or "enclose your response in double quotation marks." This dataset is used to test the ability of large language models to accurately follow given instructions.

\subsection{HELM Evaluation}
HELM\cite{Liang2022HolisticEO} is a holistic evaluation of Language Models (LMs) to improve the transparency of language models, providing a more comprehensive understanding of the capabilities, risks, and limitations of language models.
Specifically, differing from other evaluation methods, HELM holds that a holistic evaluation of language models should focus on the following three factors:

\paragraph{(1) Broad coverage.} During the development, language models can be adapted to various NLP tasks (e.g., sequence labeling and question answering), thus, the evaluation of language models needs to be carried out in a wide range of scenarios. To involve all potential scenarios, HELM proposed a top-down taxonomy, which begins by compiling all existing tasks in a major NLP conference (ACL2022) into a task space and dividing each task into the form of scenarios (e.g., languages) and metrics (e.g., accuracy). Then when facing a specific task, the taxonomy would select one or more scenarios and metrics in the task space to cover it. By analyzing the structure of each task, HELM clarifies the evaluation content (task scenarios and metrics) and improves the scenario coverage of language models from 17.9\% to 96.0\%.

\paragraph{(2) Multi-metric measurement.} In order to enable human to weigh language models from different perspectives, HELM proposes multi-metric measurement. HELM has covered 16 different scenarios and 7 metrics. To ensure the results of intensive multi-metric measurement, HELM measured 98 of 112 possible core scenarios (87.5\%).

\paragraph{(3) Standardization.} The increase in the scale and training complexity of language models has seriously hindered human's understanding of the structure of each language model. To establish a unified understanding of existing language models, HELM benchmarks 30 well-known language models, covering such institutions as Google (UL2\cite{tay2022ul2}), OpenAI (GPT-3\cite{Brown2020LanguageMA}), and EleutherAI (GPT-NeoX\cite{Black2022GPTNeoX20BAO}). Interestingly, HELM pointed out that LMs such as T5 \cite{Raffel2019ExploringTL} and Anthropic-LMv4-s3 \cite{bai2022training} had not been directly compared in the initial work, while LLMs such as GPT-3 and YaLM were still different from their corresponding reports after multiple evaluations.

\subsection{LLM As a Judge}

LLM as a judge refers to a set of methods that utilize powerful LLMs, particularly GPT-4 \cite{OpenAI2023GPT4TR}, to evaluate the outputs of other LLMs.
There are three primary reasons for this approach: (1) \textbf{Efficiency} – Manually reviewing numerous LLM outputs can be labor-intensive, whereas GPT-4 can evaluate large-scale responses quickly, saving both time and effort; (2) \textbf{Reliable Benchmark} – As one of the most advanced models available, GPT-4 provides a dependable benchmark, allowing researchers to compare the performance of different LLMs against a high standard; and (3) \textbf{Enhanced Capability} – With improved comprehension and reasoning over previous models, GPT-4 is better suited to analyze subtle aspects of language generation and handle complex outputs from other LLMs.
In the following, we detail 4 commonly accepted judge benchmarks:

\paragraph{(1) AlpacaEval.}
AlpacaEval \cite{alpaca_eval} is an automated evaluation metric leveraging LLMs, consisting of 805 instructions selected to reflect typical user interactions from the Alpaca web demo\footnote{\url{https://crfm.stanford.edu/2023/03/13/alpaca.html}}. Specifically, for each instruction, both a baseline model $b$ (currently GPT-4 turbo \cite{OpenAI2023GPT4TR}) and the model under evaluation $m$ generate responses. A GPT-4 turbo-based evaluator then conducts a head-to-head comparison of these responses, determining the probability of favoring the evaluated model. The win rate is calculated as the expected probability that the evaluator prefers the evaluated model's response across the 805 instructions, serving as a key metric for assessing the performance of the evaluated LM chatbot.

\paragraph{(2) Length-Controlled AlpacaEval.}
Length-Controlled AlpacaEval \cite{dubois2024length} is a variation of the AlpacaEval \cite{alpaca_eval} evaluation metric, designed to minimize length bias, as the original AlpacaEval tends to favor models that produce longer responses. To achieve this goal, \citet{dubois2024length} first fit a generalized linear model to predict the annotator's (GPT-4's) preference based on three factors: (1) the instruction, (2) the model used, and (3) the length difference between the baseline and the model’s output. Then, by conditioning the length difference to 0, \citet{dubois2024length} can obtain the length-controlled preference. This idea, which predicts the outcome while conditioning on the length difference (mediator), is a common technique in statistical inference, and by introducing it, Length-Controlled AlpacaEval increases the Spearman correlation with LMSYS’ Chatbot Arena from 0.94 to 0.98.

\paragraph{(3) MT-Bench.}
Currently, close-ended evaluations only measure LLMs’ core capability on a confined set of tasks, such as MMLU \cite{hendrycks2020measuring} for multi-choice decisions, without adequately assessing its alignment with human preference in open-ended tasks, such as the ability to adhere to instructions in multi-turn dialogues accurately. To alleviate this issue, \citet{zheng2023judging} introduced MT-Bench, which comprises 80 high-quality multi-turn questions designed to assess LLMs' capability in multi-turn conversations and instruction-following, with evaluations conducted using GPT-4. MT-Bench is meticulously crafted to cover eight common tasks: writing, roleplay, extraction, reasoning, math, coding, knowledge I (STEM), and knowledge II (humanities/social sciences). For alignment, GPT-4 achieves over 80\% agreement, comparable to the level of agreement among humans, making it a more reliable choice for a public benchmark.

\paragraph{(4) WildBench.}
Although the above evaluations are effective, they have notable limitations in task composition and skill coverage. For example, MT-Bench \cite{hendrycks2020measuring} includes only 80 test instructions, while AlpacaEval \cite{alpaca_eval} features many straightforward tasks, such as “What is the capital of Australia?”
To address this issue, \citet{lin2024wildbench} introduced WildBench, comprising 1,024 test instructions carefully curated from extensive human-chatbot conversation logs. WildBench draws directly from real-world user interactions, featuring numerous challenging tasks, such as coding and math problem-solving. These tasks frequently demand critical thinking, making WildBench significantly more difficult than other benchmarks. 
WildBench utilizes two metrics: WB-Reward for pairwise comparisons and WB-Score for individual assessments. Both metrics show strong alignment with human evaluations, with Pearson correlations of 0.98 for WB-Reward and 0.95 for WB-Score when compared to the human-voted ratings.

\subsection{Low-resource Instruction Tuning}
\citet{gupta2023instruction} attempts to estimate the minimal downstream training data required by SFT models to match the SOTA supervised models over various tasks.
\citet{gupta2023instruction} conducted experiments on 119 tasks from Super Natural Instructions (SuperNI) in both single-task learning (STL) and multi-task learning (MTL) settings. The results indicate that in the STL setting, SFT models with only 25\% of downstream training data outperform the SOTA models on those tasks, while in the MTL setting, just 6\% of downstream training data can lead SFT models to achieve the SOTA performance. These findings suggest that instruction
tuning can effectively assist a model in quickly learning a task even with limited data. 

However, due to resource limitations, \citet{gupta2023instruction} did not conduct experiments on LLMs, like T5-11B. So, to gain a more comprehensive understanding of the SFT models, further investigation using larger language models and datasets is necessary.

\subsection{Smaller Instruction Dataset}
SFT  requires a substantial amount of specialized instruction data for training. \citet{Zhou2023LIMALI} hypothesized  that the pre-trained LLM only has to learn the style or format to interact with users and proposed LIMA that achieves  strong performance by  fine-tuning an LLM on only 1,000 carefully selected training examples.

Specifically, LIMA first manually curates 1,000 demonstrations with high-quality prompts and responses. Then the 1,000 demonstrations are used to fine-tune the pre-trained 65B-parameter LLaMa  \cite{touvron2023llama}. By comparison, across more than 300 challenging tasks, LIMA outperfrms GPT-davinci003 \cite{Brown2020LanguageMA}, which was fine-tuned on 5,200 examples by human feedback tuning. Moreover, with only half amount of demonstrations, LIMA achieves equivalent results to GPT-4 \cite{OpenAI2023GPT4TR}, Claude \cite{bai2022constitutional}, and Bard\footnote{Bard, designed by Google, is an interface to generative AI platform, and the link is: https://ai.google/static/documents/google-about-bard.pdf}.
Above all, LIMA demonstrated that LLMs' powerful knowledge and capabilities can be exposed to users with only a few carefully curated instructions to fine-tune. 

\subsection{Evaluating  Instruction Tuning Datasets}
The performance of SFT model highly depends on the SFT datasets. 
However, 
there lacks of evaluations for these SFT datasets from open-ended and subjective aspects.

To address this issue, \citet{Wang2023HowFC} performs dataset evaluation by 
fine-tuning the LLaMa model \cite{touvron2023llama} on various of open SFT datasets and measure different fine-tuned models through both automatic and human evaluations. 
An additional model is trained on the combination of SFT datasets. 
For the results, \citet{Wang2023HowFC} showed that there is not a single best SFT dataset across all tasks, while by manually combining datasets it can achieve the best overall performance. Besides, \citet{Wang2023HowFC} pointed out that though SFT can bring large benefits on LLMs at all sizes, smaller models and models with a high base quality benefit most from SFT. For human evaluations, \citet{Wang2023HowFC} a larger model is more likely to gain a higher acceptability score.

\subsection{Proprietary LLMs Imitation}
LLMs imitation is an approach that collects outputs from a stronger model, such as a proprietary system like ChatGPT, and uses these outputs to fine-tune an open-source LLM. Through this way, an open-source LLM may get competitive capabilities with any proprietary model.

\citet{gudibande2023false} conducted several experiments to critically analyze the efficacy of model imitation. Specifically, \citet{gudibande2023false} first collected datasets from outputs of ChatGPT over broad tasks. Then these datasets were used to fine-tune a range of models covering sizes from 1.5B to 13B, base models GPT-2 and LLaMA, and data amounts from 0.3M tokens to 150M tokens.

For evaluations, \citet{gudibande2023false} demonstrated that on tasks with supported datasets, imitation models are far better than before, and their outputs appear similar to ChatGPT's. While on tasks without imitation datasets, imitation models do not have improvement or even decline in accuracy. 

Thus, \citet{gudibande2023false} pointed out that it's the phenomenon that imitation models are adept at mimicking ChatGPT's style (e.g., being fluent, confident and well-structured) that makes researchers have the illusion about general abilities of imitation models. So, \citet{gudibande2023false} suggested that instead of imitating proprietary models, researchers had better focus on improving the quality of base models and instruction examples.

\section{The Role of Instruction Fine-tuning}
\label{sec:the_role_of_instruction_fine_tuning}

Instruction fine-tuning (IF), also known as supervised fine-tuning (SFT), is a conventional alignment approach that trains models on example prompts paired with corresponding responses to ensure the model's output aligns with user instructions and intended goals.
More recently, some reinforcement learning (RL) based methods \cite{wang2024reinforcement}, e.g., reinforcement learning from human feedback (RLHF) \cite{ouyang2022training}, direct preference optimization (DPO) \cite{rafailov2023direct}, and group relative policy optimization (GRPO) \cite{shao2024deepseekmath}, and various prompt engineering strategies have emerged as alternatives or complements.
Thus, in this section, we will review each method's role in aligning LLMs, and examine whether SFT remains necessary.
Further more, we also consider the risk of superficial alignment, i.e. alignment that changes only the model's surface behavior (tone, style) without imparting deeper understanding.

\subsection{SFT Compared with Other Alignment Methods}
Below, we begin by outlining three widely used alignment approaches, RLHF, DPO, and prompt engineering, highlighting their strengths and weaknesses.
Then, we explain why SFT remains an essential component of contemporary alignment pipelines.

\subsubsection{Reinforcement Learning from Human Feedback (RLHF)}

RLHF is the dominant alignment paradigm, and typically proceeds in three phases: (1) supervised fine-tuning (SFT) on human large amounts of instruction-answer pairs, (2) training a reward model on human-ranked responses, and (3) using policy optimization (e.g. PPO \cite{schulman2017proximal}) to maximize the reward model's feedback \cite{chen2025extracting}. 
This pipeline can deeply adjust model behavior to complex user preferences. RLHF has enabled remarkable capabilities (e.g. nuanced help, factuality), but at high cost and complexity. 
It requires extensive human data, careful RL tuning, and often suffers stability issues and ``reward hacking'' (the model finds loopholes in the reward model) \cite{xiao2024comprehensive,wang2024reinforcement}. 
Because RLHF optimization is resource-intensive and sensitive to hyper parameters, simpler alternatives have been sought.

\paragraph{Advantages.}
The key strength of RLHF lies in its ability to guide models toward high-level objectives, such as helpfulness and safety, that are not explicitly encoded in the training data, demonstrating strong empirical performance in aligning models with user intent when well-tuned \cite{wang2024reinforcement,chen2025extracting}.

\paragraph{Limitations.}
RLHF's complexity is a downside. It typically requires starting from an SFT-trained model, i.e., a model that already follows instructions to some degree, because training RL from a raw base model is difficult \cite{trivedi2025align}. 
The multi-stage pipeline (SFT, reward model, and PPO) is time-consuming and brittle \cite{chen2025extracting}. 
In practice, researchers and practitioners often still perform an initial instruction fine-tuning, even when using RLHF, to establish a reasonable base policy. 
Moreover, RLHF can introduce ``alignment tax'' (performance drop on some tasks) and can fail to generalize if the reward model is mis-specified \cite{xiao2024comprehensive}.

\subsubsection{Direct Preference Optimization (DPO)}

Direct Preference Optimization (DPO) \cite{rafailov2023direct} is a recently proposed RL-free alignment method that directly fine-tunes on preference pairs. 
Instead of learning a separate reward model and running RL, DPO casts alignment as a supervised objective: for each prompt and pair of outputs (preferred vs. dispreferred), it adjusts the model's logits to increase the probability of the preferred output. DPO’s loss is equivalent to a Bradley–Terry pairwise classification (a logit-ratio objective) to bypass policy-gradient RL entirely.

\paragraph{Advantages.}

Because DPO fits into a standard maximum-likelihood fine-tuning framework, it is far simpler and more stable than PPO-based RLHF \cite{xu2024dpo,xiao2024comprehensive,wang2024reinforcement}
. Studies report that DPO matches or exceeds RLHF performance on tasks like summarization or helpfulness with fewer preference examples. 
Compared to RLHF, DPO has been shown to be stable, performant, and computationally lightweight in various applications. 
It does not require expensive RL infrastructure or tuning of PPO hyper-parameters, making it reproducible and easier to deploy. 
Practitioners, e.g. OpenAI \footnote{\url{https://cookbook.openai.com/examples/fine_tuning_direct_preference_optimization_guide}} and DeepSeek \cite{shao2024deepseekmath} note that starting DPO from an already fine-tuned model improves results, using SFT to establish a robust initial policy.

\paragraph{Limitations.}

DPO still depends on quality preference data, and like any offline method it can suffer if the data distribution shifts. Recent analyses have identified issues with DPO: because it employs an implicit reward tied to the policy, it can bias the model toward out-of-distribution outputs and even degrade generalization \cite{xiao2024comprehensive,wang2024reinforcement}. 
Variants, e.g. KL-constrained or semantics-aware DPO, are being developed to mitigate these problems, but it remains true that DPO typically benefits from starting with a good initial model. In practice, most implementations still use an SFT-tuned model before running DPO, echoing the RLHF pipeline. Thus, while DPO simplifies alignment, it has not eliminated the need for supervised tuning in many cases.

\subsubsection{Prompt Engineering (In-Context Learning)}

Prompt engineering aligns model behavior without any fine-tuning. Instead, it leverages the model's existing capabilities by crafting prompts, including instructions, few-shot examples, or chain-of-thought cues, to elicit desired outputs. 
Recent work treats prompt design itself as an optimization problem: one can optimize a prompt string or learn soft prompts to maximize human-aligned metrics \cite{trivedi2025align}. 
Importantly, prompt-based methods assume no weight updates, which means that they work without any post training.

\paragraph{Advantages.}

The biggest benefit is that no retraining is required. Prompt optimization can effectively align LLMs even when parameter fine-tuning is not feasible \cite{trivedi2025align}. 
This is appealing for large models with fixed parameters (APIs or frozen on-device models). Some experiments have shown that with well-designed prompts, base LLMs can achieve high performance on instruction-following tasks. For example, \citet{lin2023urial} introduced URIAL, a method that uses only a few stylistic examples in-context (plus a system prompt) to steer the model, and found it matched or even surpassed fully-tuned models in many benchmarks. 
This suggests that clever prompt engineering alone can yield strong alignment in some cases \cite{wang2023gpt,sun2023sentiment,sun2023pushing,wang2023sim,sun2023text}. 
Prompting has obvious speed and convenience advantages: it requires no training data or compute, and can be iterated quickly by users.

\paragraph{Limitations.}

Prompt-based alignment has inherent limits. The model's context window bounds how much instruction or example content can be provided so that very complex tasks may simply not fit. 
More importantly, prompt methods generally induce superficial compliance rather than truly altering the model's knowledge. They leverage already-encoded patterns in the model, but cannot add new capabilities or correct deep misunderstandings. 
In practice, prompt engineering often produces brittle behaviors: slight rephrasing can break performance, and malicious users can ``jailbreak'' around prompts to elicit bad outputs. For instance, \citet{chen2025unleashing} note that although prompt-based ICL can align a model to some extent, it does so mainly by inserting stylistic cues and does not fundamentally change the model's reasoning process. 
In short, prompt engineering can quickly achieve surface-level alignment (tone, disclaimers, formatting), but cannot replace weight tuning for deep or novel tasks, and for complex reasoning, mathematics, or new knowledge integration typically require additional fine-tuning.

\subsubsection{The Continued Necessity of SFT}

Given these techniques, a key question is whether instruction fine-tuning (IF) or supervised fine-tuning (SFT) remains necessary in modern pipelines. Empirical evidence suggests it does. Both RLHF and DPO pipelines almost universally incorporate an initial SFT stage. 
In RLHF this is explicit, which SFT usually serves as the first phase. 
For DPO, while the final optimization is simpler, practitioners generally first fine-tune on good example responses to establish a robust initial policy, which stabilizes subsequent DPO refinement. 
The OpenAI alignment guide explicitly recommends performing supervised fine-tuning on a subset of preferred responses before DPO to improve alignment and convergence\footnote{\url{https://cookbook.openai.com/examples/fine_tuning_direct_preference_optimization_guide}}. 
In other words, even with DPO, a round of instruction tuning yields better outcomes. 
On the other hand, prompt-based methods show that in principle one can align some models without any fine-tuning \cite{wang2023gpt,sun2023sentiment,sun2023pushing,wang2023sim,sun2023text}. 
Other work \cite{lin2023unlocking} demonstrates that you can in effect ``unlock'' base LLMs by providing few-shot prompts, achieving performance close to tuned models without SFT. 
However, these tuning-free methods tend to rely on pre-existing capabilities. If a base model genuinely lacks a skill, such as solving a type of math problem it was never pre-trained on, no prompt will fix it, and only updating weights can. Moreover, some recent studies \cite{parthasarathy2024ultimate} find that for reasoning and knowledge tasks, performance continues to scale with more fine-tuning data, suggesting base models do improve under instruction tuning. 
In summary, prompt methods can sometimes obviate SFT for shallow alignment, but for robust alignment pipelines or domain specific alignment (e.g., Medicinal Chemistry), supervised fine-tuning is still regarded as essential groundwork. New research even explores hybrid tricks, such as ``instruction residuals'' from an older model added to a new base, to avoid re-training, but these rely on existing tuned models as sources. The prevailing practice remains: use SFT to teach the model the format and style of responses, then refine preferences via RLHF or DPO.

\subsection{Superficial Alignment}

Despite the impressive improvements in the performance of instruction tuning, there lacks clarity about the specific knowledge that models acquire through instruction tuning, raising questions about:
\textit{Does instruction tuning just learn Pattern Copying?} or \textit{How exactly does the alignment tuning transform a base LLM?}

To answer these questions, 
\citet{Kung2023DoMR} delves into the analysis of how models make use of instructions during SFT by comparing the tuning when provided with altered instructions versus the original instructions.

Specifically, \citet{Kung2023DoMR} creates simplified task definitions that remove all semantic components, leaving only the output information. In addition, \citet{Kung2023DoMR} also incorporates delusive examples that contain incorrect input-output mapping. Surprisingly, the experiments show that models trained on these simplified task definitions or delusive examples can achieve comparable performance to the ones trained on the original instructions and examples. Moreover, the paper also introduces a baseline for the classification task with zero-shot, which achieves similar performance to SFT in low-resource settings.

Similar to the findings of \citet{Kung2023DoMR}, several subsequent studies \cite{Zhou2023LIMALI, lin2023unlocking} reached the same conclusion: the observed performance improvements in current SFT models are often due to superficial alignment. This means the models excel at recognizing superficial alignment, such as mastering output formats and making educated guesses, rather than truly understanding and learning the underlying tasks.

\section{Conclusion}
This work surveys recent advances in the fast growing field of instruction tuning, which can also be referred to as supervised fine-tuning (SFT). 
We make a systematic review of the literature, including the general methodology of SFT, 
the construction of SFT datasets, the training of SFT models, 
SFT's applications to different modalities, domains and application. 
We also review analysis on SFT models to discover both their advantages and potential pitfalls. 
We hope this work will 
act as a stimulus to motivate further endeavors to address the deficiencies of current SFT models.

\bibliographystyle{acl_natbib}
 
\bibliography{custom}

\appendix

\section{Datasets} 
\label{appendix:datasets}

\begin{table*}[t]
\centering
\small
\begin{adjustbox}{max width=0.9\textwidth}
\begin{threeparttable}
\begin{tabular}{p{3.8cm}|lccccc}
\toprule 
{\bf Type} & {\bf Dataset Name} & {\bf \# of Instances} & {\bf \# of Lang} & {\bf Construction} & {\bf Open-source} \\\midrule

\multirow{13}{*}{\bf Human-Crafted} & UnifiedQA~\citep{khashabi2020unifiedqa}\tnotex{id:1} & 750K & En & human-crafted & Yes\\
& UnifiedSKG~\citep{Xie2022UnifiedSKGUA}\tnotex{id:3} & 0.8M & En & human-crafted & Yes\\
& Natural Instructions~\citep{honovich2022unnatural}\tnotex{id:4} & 193K & En  & human-crafted & Yes \\
& Super-Natural Instructions~\citep{supernaturalinstructions}\tnotex{id:5} & 5M & 55 Lang & human-crafted & Yes \\
& P3~\citep{sanh2021multitask}\tnotex{id:6} & 12M & En & human-crafted & Yes \\
& xP3~\citep{muennighoff2022crosslingual}\tnotex{id:7} & 81M & 46 Lang & human-crafted & Yes \\
& Flan 2021~\citep{longpre2023flan}\tnotex{id:8} & 4.4M & En & human-crafted & Yes \\
& COIG~\citep{Zhang2023ChineseOI}\tnotex{id:9} & - & - & - & Yes \\
&  InstructGPT~\citep{ouyang2022training} & 13K & Multi & human-crafted & No \\
& Dolly~\citep{conover2023free}\tnotex{id:22} & 15K & En & human-crafted & Yes \\
& LIMA~\citep{Zhou2023LIMALI}\tnotex{id:18} & 1K & En & human-crafted & Yes \\
& ChatGPT~\citep{chatgpt} & - & Multi & human-crafted & No \\
& OpenAssistant~\citep{kopf2023openassistant}\tnotex{id:17} & 161,443 & Multi & human-crafted & Yes \\
\midrule 


\multirow{20}{*}{\shortstack{\bf Synthetic Data \\ \bf (Distillation)}} & OIG~\citep{2023oig}\tnotex{id:2} & 43M & En & ChatGPT (No technique reports) & Yes \\
& Unnatural Instructions~\citep{honovich2022unnatural}\tnotex{id:10} & 240K & En & InstructGPT-Generated & Yes \\
& InstructWild~\citep{instructionwild}\tnotex{id:12} & 104K & - & ChatGPT-Generated & Yes\\
& Evol-Instruct / WizardLM~\citep{xu2023wizardlm}\tnotex{id:13} & 52K & En & ChatGPT-generated & Yes \\
& Alpaca~\citep{taori2023alpaca}\tnotex{id:14}  & 52K & En & InstructGPT-generated & Yes \\
& LogiCoT~\citep{Liu2023LogiCoTLC}\tnotex{id:15} & - & En & GPT-4-Generated & Yes \\
& GPT-4-LLM~\citep{peng2023instruction}\tnotex{id:20} & 52K & En\&Zh & GPT-4-Generated & Yes \\
& Vicuna~\citep{chiang2023vicuna} & 70K & En & Real User-ChatGPT Conversations & No \\
& Baize v1~\citep{DatabricksBlog2023DollyV2}\tnotex{id:21} & 111.5K & En & ChatGPT-Generated & Yes \\
& UltraChat~\citep{ding2023enhancing}\tnotex{id:16} & 675K & En\&Zh & GPT 3/4-Generated & Yes \\
& Guanaco~\citep{Guanaco}\tnotex{id:19} & 534,530 & Multi & GPT (Unknown Version)-Generated & Yes \\
& Orca~\citep{mukherjee2023orca}\tnotex{id:23} & ~1.5M & En & GPT 3.5/4-Generated & Yes \\
& ShareGPT\tnotex{id:24} & 90K & Multi & Real User-ChatGPT Conversations & Yes \\
& WildChat\tnotex{id:25} & 150K & Multi & Real User-ChatGPT Conversations & Yes \\
& WizardCoder~\citep{luo2023wizardcoder} & - & Code & LLaMa 2-Generated & No \\
& Magicoder~\citep{wei2023magicoder}\tnotex{id:26} & 75K/110K & Code & GPT-3.5-Generated & Yes \\
& WaveCoder~\citep{yu2023wavecoder} & - & Code & GPT 4-Generated & No \\
& Phi-1~\citep{gunasekar2023textbooks}\tnotex{id:27} & 6B Tokens & Code Q and A & GPT-3.5-Generated & Yes \\
& Phi-1.5~\citep{li2023textbooks} & - & Code Q and A & GPT-3.5-Generated & No \\
& Nectar~\citep{zhu2023starling}\tnotex{id:28} & ~183K & En & GPT 4-Generated & Yes \\

\midrule 

\multirow{3}{*}{\shortstack{\bf Synthetic Data \\ \bf (Self-Improvement)}} & Self-Instruct~\citep{wang2022self}\tnotex{id:11} & 52K & En & InstructGPT-Generated & Yes \\
& Instruction Backtranslation~\citep{li2023self} & 502K  & En & LLaMa-Generated & No \\
& SPIN~\citep{chen2024self}\tnotex{id:29} & 49.8K & En & Zephyr-Generated & Yes \\

\midrule 

\multirow{3}{*}{\shortstack{\bf Reasoning Data}} & PRM800K~\citep{wang2022self}\tnotex{id:30} & 800K & Math & human-crafted \& GPT-Generated & Yes \\
& O1-Journey~\citep{li2023self}\tnotex{id:31} & 677  & Math & human-crafted \& GPT-Generated & Yes \\
& Self-Explore~\citep{chen2024self} & - & Math & GPT-Generated & No \\
& MARIO~\citep{chen2024self}\tnotex{id:32} & 28.8K & Math & human-crafted \& GPT-Generated & Yes \\
& MathGenie~\citep{chen2024self} & 170K & Math & GPT-Generated & No \\
& DeepSeekMath~\citep{chen2024self}\tnotex{id:33} & 120B & Math & human-crafted \& GPT/DeepSeek-Generated & Yes \\
& Compute-Optimal Sampling~\citep{chen2024self} & - & Math & GPT-Generated & No \\
& MathScale~\citep{chen2024self}\tnotex{id:34} & 2M & Math & GPT-Generated & Yes \\
& G-LLaVA~\citep{chen2024self}\tnotex{id:35} & 170K & Math & GPT-Generated & Yes \\

\bottomrule
\end{tabular}
\end{threeparttable}
\end{adjustbox}
\begin{multicols}{2}
\begin{tablenotes}
\item[1] \label{id:1} {$^1$ https://github.com/allenai/unifiedqa} 
\item[2] \label{id:2} {$^2$ https://github.com/LAION-AI/Open-Instruction-Generalist}
\item[3] \label{id:3} {$^3$ https://github.com/hkunlp/unifiedskg}
\item[4] \label{id:4} {$^4$ https://github.com/allenai/natural-instructions-v1}
\item[5] \label{id:5} {$^5$ https://github.com/allenai/natural-instructions}
\item[6] \label{id:6} {$^6$ https://huggingface.co/datasets/bigscience/P3} 
\item[7] \label{id:7} {$^7$ https://github.com/bigscience-workshop/xmtf}
\item[8] \label{id:8} {$^8$ https://github.com/google-research/FLAN}
\item[9] \label{id:9} {$^9$ https://github.com/BAAI-Zlab/COIG}
\item[10] \label{id:10} {$^{10}$ https://github.com/orhonovich/unnatural-instructions}
\item[11] \label{id:11} {$^{11}$ https://github.com/yizhongw/self-instruct}
\item[12] \label{id:12} {$^{12}$ https://github.com/XueFuzhao/InstructionWild}
\item[13] \label{id:13} {$^{13}$ https://github.com/nlpxucan/evol-instruct} 
\item[14] \label{id:14} {$^{14}$ https://github.com/tatsu-lab/stanford\_alpaca}
\item[15] \label{id:15} {$^{15}$  https://github.com/csitfun/LogiCoT}
\item[16] \label{id:16} {$^{16}$ https://github.com/thunlp/UltraChat\#data}
\item[17] \label{id:17} {$^{17}$ https://github.com/LAION-AI/Open-Assistant}
\item[18] \label{id:18} {$^{18}$ https://huggingface.co/datasets/GAIR/lima}
\item[19] \label{id:19} {$^{19}$ https://huggingface.co/datasets/JosephusCheung/GuanacoDataset}
\item[20] \label{id:20} {$^{20}$ https://github.com/Instruction-Tuning-with-GPT-4/GPT-4-LLM}
\item[21] \label{id:21} {$^{21}$ https://github.com/project-baize/baize-chatbot}
\item[22] \label{id:22} {$^{22}$ https://huggingface.co/datasets/databricks/databricks-dolly-15k}
\item[23] \label{id:23} {$^{23}$ https://huggingface.co/datasets/Open-Orca/OpenOrca}
\item[24] \label{id:24} {$^{24}$ https://huggingface.co/datasets/RyokoAI/ShareGPT52K}
\item[25] \label{id:25} {$^{25}$ https://huggingface.co/datasets/allenai/WildChat}
\item[26] \label{id:26} {$^{26}$ https://github.com/ise-uiuc/magicoder?tab=readme-ov-file\#-dataset}
\item[27] \label{id:27} {$^{27}$ https://huggingface.co/microsoft/phi-1}
\item[28] \label{id:28} {$^{28}$ https://huggingface.co/datasets/berkeley-nest/Nectar}
\item[29] \label{id:29} {$^{29}$ https://github.com/uclaml/SPIN?tab=readme-ov-file\#Data}
\item[30] \label{id:30} {$^{30}$ https://github.com/openai/prm800k}
\item[31] \label{id:31} {$^{31}$ https://github.com/GAIR-NLP/O1-Journey}
\item[32] \label{id:32} {$^{32}$ https://github.com/MARIO-Math-Reasoning/MARIO}
\item[33] \label{id:33} {$^{33}$ https://github.com/deepseek-ai/DeepSeek-Math}
\item[34] \label{id:34} {$^{34}$ https://github.com/XylonFu/MathScale}
\item[35] \label{id:35} {$^{35}$ https://github.com/pipilurj/G-LLaVA}
\end{tablenotes}
\end{multicols}
\caption{An overview of instruction tuning datasets.}
\label{tab:llms_traindata_table}
\end{table*}

Table \ref{tab:llms_traindata_table} gives an overview of our collected datasets.

\label{sec:appendix}

\end{document}